%% file: main.tex
\documentclass[10pt,twocolumn,letterpaper]{article}

\usepackage[pagenumbers]{cvpr} 


\definecolor{cvprblue}{rgb}{0.21,0.49,0.74}
\usepackage[accsupp]{axessibility}
\usepackage[pagebackref,breaklinks,colorlinks,allcolors=cvprblue]{hyperref}
\usepackage{dashrule} 
\usepackage{tikz}
\usepackage{calc}
\usepackage{xcolor}
\usepackage{algorithm}
\usepackage{algpseudocode}

\def\paperID{15932} 
\def\confName{CVPR}
\def\confYear{2025}

\title{DTGBrepGen: A Novel B-rep Generative Model through Decoupling Topology and Geometry}

\author{
Jing Li\textsuperscript{1} \quad
Yihang Fu\textsuperscript{1} \quad
Falai Chen\textsuperscript{1}\thanks{Corresponding author} \\
\textsuperscript{1} University of Science and Technology of China
}

\begin{document}
\maketitle

\begin{abstract}
\hypersetup{colorlinks=true, urlcolor=red!70!white}
Boundary representation (B-rep) of geometric models is a fundamental format in Computer-Aided Design (CAD). However, automatically generating valid and high-quality B-rep models remains challenging due to the complex interdependence between the topology and geometry of the models. Existing methods tend to prioritize geometric representation while giving insufficient attention to topological constraints, making it difficult to maintain structural validity and geometric accuracy. In this paper, we propose DTGBrepGen, a novel topology-geometry decoupled framework for B-rep generation that explicitly addresses both aspects. Our approach first generates valid topological structures through a two-stage process that independently models edge-face and edge-vertex adjacency relationships. Subsequently, we employ Transformer-based diffusion models for sequential geometry generation, progressively generating vertex coordinates, followed by edge geometries and face geometries which are represented as B-splines. Extensive experiments on diverse CAD datasets show that DTGBrepGen significantly outperforms existing methods in both topological validity and geometric accuracy, achieving higher validity rates and producing more diverse and realistic B-reps. Our code is publicly available at \href{https://github.com/jinli99/DTGBrepGen}{https://github.com/jinli99/DTGBrepGen.}
\end{abstract}   

\section{Introduction}
\label{sec:intro}
Boundary representation (B-rep) \cite{weiler1986topological} is the predominant format for 3D shape modeling in Computer-Aided Design (CAD). Unlike mesh-based representations, which rely on planar facets and linear edges, B-rep represents objects using parametric surfaces (faces), parametric curves (edges), and vertices to define both topology and geometry. This detailed structure enables B-rep models to be able to accurately capture complex surfaces and solids, making them indispensable in engineering and manufacturing.

However, automatically generating B-reps poses significant challenges due to the need to accurately model both topological relationships and geometric representations. Many existing CAD generation methods address this challenge indirectly by generating sequences of modeling commands ~\cite{wu2021deepcad, xu2022skexgen, xu2023hierarchical, li2023secad}, from which the B-rep can be recovered in postprocess using a solid modeling kernel. However, these methods are constrained by limited datasets ~\cite{willis2021fusion, wu2021deepcad} and are typically confined to basic operations like sketching and extruding, making them suitable only for simpler shapes. Among current B-rep generative models, SolidGen ~\cite{jayaraman2023solidgen} and BrepGen ~\cite{xu2024brepgen} have made strides in direct B-rep generation. However, these approaches predominantly emphasize on geometric attributes while neglecting the automatic generation of topological structures. As a result, SolidGen is restricted to simpler prismatic shapes and lacks support for more complex surfaces, while BrepGen struggles with reliably reconstructing topology from its generated geometric attributes. These limitations highlight the need for more robust approaches capable of handling both topological validity and geometric complexity in B-reps.

For this purpose, we present DTGBrepGen, a novel approach to B-rep generation that decouples topology from geometry, addressing the challenges of modeling both aspects simultaneously. DTGBrepGen first generates a valid topological structure, which defines the connectivity between faces, edges, and vertices. This step is particularly challenging due to strict topological constraints—such as each edge must connect two vertices, and edges on each face must form closed loops. To address this, we design a two-phase topology generation process with Transformer encoder-decoder architectures: the first phase generates edge-face adjacencies, followed by the second phase establishing edge-vertex connections. With this established topology, we employ Transformer-based diffusion models \cite{vaswani2017attention, ho2020denoising, dhariwal2021diffusion} to generate geometric attributes. In contrast to existing methods like UV-Net \cite{jayaraman2021uv} and BrepGen \cite{xu2024brepgen} that rely on discrete point sampling, we employs B-spline representations for curves and surfaces. This enables direct learning of control point distributions, resulting in more accurate and mathematically precise geometric representations. In summary, the contributions of this paper are as follows: 
\begin{itemize} 
\item We introduce DTGBrepGen, a novel framework for B-rep generation that explicitly addresses both topology and geometry separately. Our key innovation is a two-phase topology generation strategy designed to effectively handle complex topological constraints. 
\item We develop an advanced geometric generation pipeline that leverages Transformer-based diffusion models and B-spline representations, resulting in more precise and compact geometric outputs.
\item Comprehensive experiments are provided to show that DTGBrepGen surpasses existing approaches in producing valid, diverse, and high-quality B-rep models. 
\end{itemize}

\section{Related work}
The generation of CAD models has emerged as a significant research area, with various approaches targeting different representation formats. This section reviews some key methodologies for CAD model generation.
\subsection{CAD model generation from point clouds}
Generating CAD models from point clouds plays a key role in reverse engineering~\cite{uy2022point2cyl, lambourne2022reconstructing, guo2022complexgen}. While traditional methods converting point clouds to meshes before CAD reconstruction~\cite{benkHo2004segmentation}, recent approaches streamline this by directly predicting CAD structures. For instance, Point2Cyl~\cite{uy2022point2cyl} formulates extrusion cylinder decomposition using neural networks, while ComplexGen~\cite{guo2022complexgen} directly generates B-rep models by identifying geometric primitives and relationships. Despite recent advances, generating accurate CAD models from point clouds remains challenging due to issues like data incompleteness, high geometric complexity, and the difficulty of inferring precise topological relationships.
\subsection{Constructive solid geometry}
Constructive Solid Geometry (CSG) creates 3D shapes by combining simple primitives like cubes and spheres using Boolean operations such as union and subtraction. This representation has been extensively used in shape programs and parametric modeling due to its simplicity and interpretability \cite{tian2018learning, ellis2019write,ritchie2023neurosymbolic}. However, CSG's reliance on predefined primitives makes it less flexible in capturing more intricate shapes. Furthermore, converting CSG models to B-rep often introduces geometric artifacts such as sliver faces, which complicates downstream tasks.
\subsection{CAD command generation}
Recent advances in CAD modeling enable sequence generation of CAD commands directly from parametric files, producing editable models. DeepCAD~\cite{wu2021deepcad} pioneered the modeling of CAD command sequences, laying a foundation for further developments~\cite{xu2022skexgen, xu2023hierarchical, li2023secad}. Despite this progress, these methods are mostly limited to basic operations (\eg, sketch and extrude) and struggle with complex commands like fillet and chamfer. Additionally, datasets containing CAD operations ~\cite{willis2021fusion, wu2021deepcad} are smaller (around 190K models) than datasets without operations ~\cite{koch2019abc} (around 1M models). While progress has been made, challenges remain in broadening operation ranges and dataset sizes. 
\subsection{B-rep generation}
B-rep models provide CAD representations by defining solids through vertices, edges, and faces, making them essential for capturing complex surface interactions in industrial CAD software. Prior works have explored B-rep tasks such as classification and segmentation~\cite{cao2020graph, willis2022joinable} and parametric surface generation~\cite{smirnov2021learning, wang2020pie}. PolyGen \cite{nash2020polygen} employs Transformers \cite{vaswani2017attention} and pointer networks \cite{vinyals2015pointer} to generate n-gon meshes, which are simplified instances of B-rep models with planar faces and linear edges. Building on this, recent methods like SolidGen \cite{jayaraman2023solidgen} extend to full B-rep structures via sequential generation, starting with vertices and conditionally predicting edges and faces. However, they remain limited to prismatic shapes. In contrast, BrepGen \cite{xu2024brepgen} represents B-reps as hierarchical tree structures, leveraging multiple diffusion models to generate geometric attributes, followed by a post-processing step to merge nodes and recover topology. Nevertheless, this method often struggles with accurately reconstructing topological structures. Thus, while these approaches advance  B-rep generation, they typically couple topological and geometric processes, making it challenging to produce B-rep models that are both topologically correct and geometrically precise.

\section{Generative framework of B-rep structures}
In this section, we will first describe the B-rep generation problem, and then outline the basic idea of our approach.

\begin{figure*}
    \centering
    \includegraphics[width=0.9\linewidth]{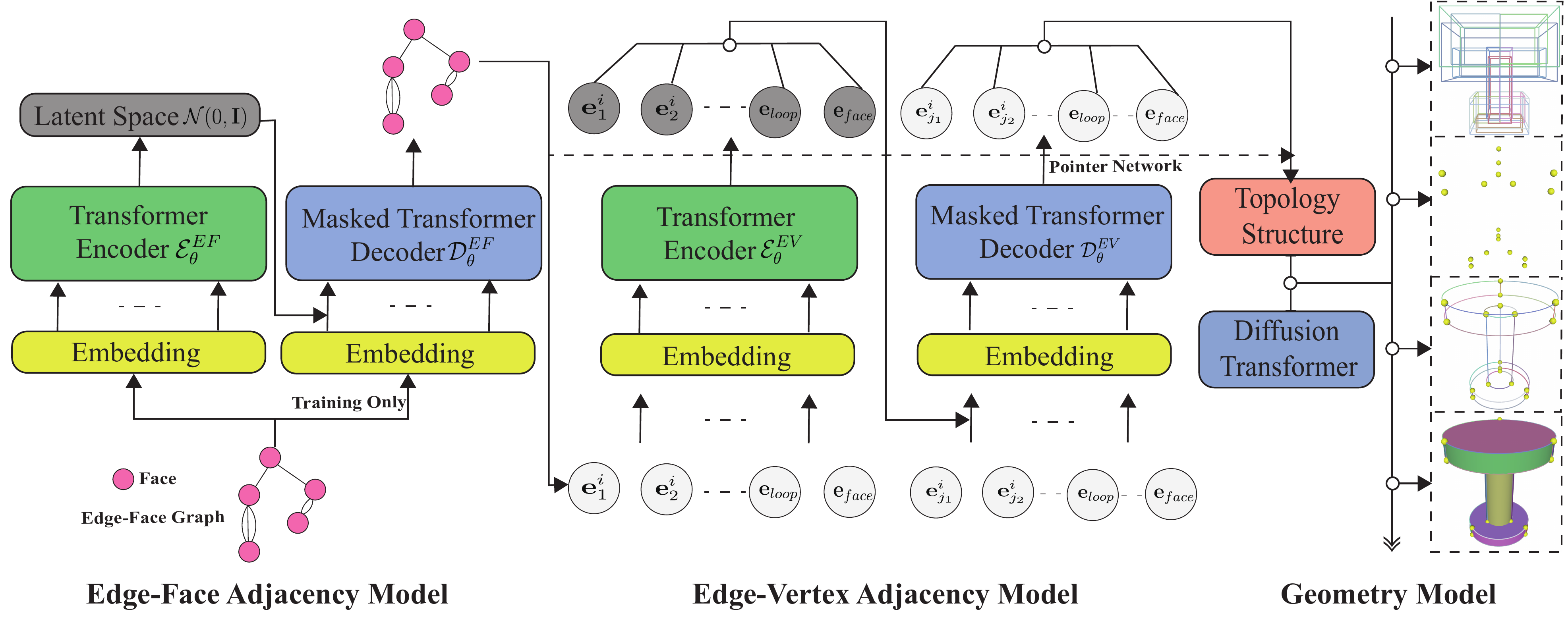}
    \caption{Overview of DTGBrepGen. The pipeline first generates a valid topological structure by two sequentially  Transformer encoder-decoder networks: one to produce edge-face adjacencies, followed by another to establish edge-vertex connections. Based on this topology, the geometry generation process employs Transformer-based diffusion models to progressively generate face bounding boxes, vertex coordinates, edge geometries, and finally face geometries.} 
    \label{fig: pipeline} 
\end{figure*}

\subsection{Problem setup}
\label{sec: problem}
Let $\mathcal{B} = \left\{\mathcal{B}_1, \mathcal{B}_2, \ldots, \mathcal{B}_N \right\}$ denote a set of real B-rep models, where each $\mathcal{B}_i = (\mathcal{T}_i, \mathcal{G}_i)$ is characterized by its topological structure $\mathcal{T}_i$ and its geometric attributes $\mathcal{G}_i$. A B-rep structure is composed of three fundamental elements: vertices, edges, and faces. The topological structure $\mathcal{T}_i$ defines the connectivity between these elements without specifying their exact positions. It describes how vertices are connected to form edges, and how edges combine to form faces. In contrast, the geometric attributes $\mathcal{G}_i$ provides the precise spatial definitions of each element. Specifically, it includes the 3D coordinates of each vertex, the curve definitions for each edge (such as lines or arcs), which are bounded by two vertices, and the surface definitions for each face (such as planes or spheres), which are enclosed by specific edges.For each $\mathcal{B}_i$, we denote the number of vertices, edges, and faces as $N_v^i$, $N_e^i$, and $N_f^i$ respectively, and the sets of vertices, edges, and faces as $\mathbf{V}_i$, $\mathbf{E}_i$, and $\mathbf{F}_i$. 

For simplicity, we limit our scope to closed B-rep models, where each edge is shared by exactly two distinct faces.  Additionally, all closed faces and edges are split along seams~\cite{jayaraman2023solidgen}, simplifying processing by ensuring each edge connects two distinct vertices. This preprocessing step allows us to more easily define and work with the topological structure. In particular, a valid topological structure must satisfy the following three conditions:
\begin{itemize}
    \item $C_1$: Each edge must be contained in exactly two distinct faces.
    \item $C_2$: Each edge must connect exactly two distinct vertices.
    \item $C_3$: The edges of each face must form closed loops.
\end{itemize}

Our goal is to develop a generative model $G_\theta$, parameterized by $\theta$, which generates new B-rep models with valid topologies and accurate geometries from a known distribution (typically standard Gaussian distribution). The generated B-rep models should approximate the underlying distribution of real B-rep models $\mathcal{B}$. In the context of existing generative frameworks \cite{DBLP:journals/corr/KingmaW13, goodfellow2014generative, ho2020denoising}, a common approach is maximum likelihood estimation, where we aim to maximize the likelihood $\prod_{i=1}^{N} P_\theta(\mathcal{B}_i)$, with $P_\theta(\mathcal{B}_i)$ denoting the probability that the model $G_\theta$ generates $\mathcal{B}_i$. This optimization ensures that the generated B-rep models closely match the true distribution of real-world B-rep structures, thereby producing realistic and valid outputs.

\subsection{Method overview}
\label{sec: overview}

Unlike previous approaches (SolidGen and BrepGen), the core of DTGBrepGen is to decouple the generation of topology from geometry, structuring the probability of generating a B-rep model $\mathcal{B}_i$ as: \begin{equation} 
P_\theta(\mathcal{B}_i) = P_\theta(\mathcal{T}_i) P_\theta(\mathcal{G}_i | \mathcal{T}_i), \label{eq: prob} 
\end{equation} 
We model each part using separate neural networks. 

We first discuss the generation of the topological structure, with the primary challenge being to ensure compliance with the three previously defined constraints. There are six possible adjacency relations between vertices, edges, and faces, derived from their pairwise combinations. Our initial step is to select a subset of these relations that meet the following criteria: 
\begin{itemize}
\item The relations should uniquely define the complete topological structure. 
\item The relations should be minimal, meaning none can be inferred from the others. 
\item The relations should be suitable for learning within a generative model. \end{itemize} 
The first two criteria ensure that the selected adjacency relations provide a “basis” for defining the topological structure. We found that the edge-face matrix $\mathbf{EF}_i \in \mathbb{N}^{N_e^i\times 2}$ and the edge-vertex matrix $\mathbf{EV}_i \in \mathbb{N}^{N_e^i\times 2}$ are such a basis for $\mathcal{T}_i$, where $\mathbf{EF}_i$ stores the IDs of the two faces connected by each edge, and $\mathbf{EV}_i$ stores the IDs of the two vertices connected by each edge in the B-rep model $\mathcal{B}_i$. Notably, when each row of $\mathbf{EF}_i$ and $\mathbf{EV}_i$ contains two distinct elements, constraints $C_1$ and $C_2$ are inherently satisfied, simplifying the topology generation process. Additionally, due to the structured representation of $\mathbf{EF}_i$ and $\mathbf{EV}_i$, they are well-suited for generative model learning. Due to the challenges of generating these matrices simultaneously, we first generate the edge-face matrix $\mathbf{EF}_i$, followed by the generation of the edge-vertex matrix $\mathbf{EV}_i$. Both generative networks employ a Transformer encoder-decoder architecture \cite{vaswani2017attention}, as illustrated in \cref{fig: pipeline}. However, not all configurations of $\mathbf{EF}_i$ and $\mathbf{EV}_i$ yield a valid topology, as they may fail to meet the constraint $C_3$, requiring edges of each face to form closed loops. This constraint is further discussed in \cref{sec: evGen}.

When the topology is established, we sequentially generate geometric attributes: vertex coordinates are generated first, followed by edge geometry, and finally face geometry. At each stage, existing topological information is integrated. For instance, when generating edge geometry, the coordinates of the two connected vertices serve as conditional inputs. Additionally, we introduce a preliminary step for generating face bounding boxes, which significantly enhances the  quality of the results. Consequently, our geometry generation architecture includes four diffusion-based generative models, each employing a Transformer encoder as the denoiser \cite{vaswani2017attention, ho2020denoising, dhariwal2021diffusion}, as shown in \cref{fig: pipeline}. We represent curves and surfaces using B-splines, enabling direct learning of  distributions of the control points. This approach provides precise geometric representations and avoids the need for discrete point sampling \cite{jayaraman2021uv} or a separate VAE \cite{DBLP:journals/corr/KingmaW13} for encoding high-dimensional point clouds~\cite{xu2024brepgen}, both of which can introduce geometric inaccuracies.

\section{Decoupling topology and geometry}
In this section, we discuss each phase of topology and geometry generation in detail.

\subsection{Edge-face adjacency generation}
\label{sec: efGen}
Directly learning the distribution of the edge-face matrices $\left\{\mathbf{EF}_i\right\}_{i=1}^{N}$ is suboptimal, as it stores face IDs which do not inherently reflect the topological structure. To address this, we introduce a more suitable and topologically equivalent representation $\left\{\mathbf{FeF}_i\right\}_{i=1}^{N}$. Here, $ \mathbf{FeF}_i \in \mathbb{N}^{N_f^i \times N_f^i}$, with each element $\mathbf{FeF}_i[k, l]$ denoting the number of edges shared between face $k$ and face $l$ in B-rep $\mathcal{B}_i$. This representation can also be viewed as a graph, as illustrated in \cref{fig: pipeline}, where nodes represent the faces of the B-rep, and edges between nodes indicate the number of shared edges between two faces. Notice that, reassigning face IDs within the same topological structure leads to different matrices $\mathbf{FeF}_i$ , even if they represent the same topology. To standardize, we assign face IDs by sorting faces in ascending order according to their edge count. For faces with identical edge counts, we assign their order randomly, as we found this randomness assignment minimal impact on training performance. Given the symmetry of the $\mathbf{FeF}_i$ matrix, we extract its upper triangular portion and convert it into a sequence: 
\begin{equation} 
\mathbf{EF}_i^{seq} := \left\{ \mathbf{FeF}_i[1, 2], \mathbf{FeF}_i[1, 3], \cdots, \mathbf{FeF}_i[N_f^i-1, N_f^i]\right\}, 
\end{equation} 

\textbf{Shared-edges embedding.} The sequence $\mathbf{EF}_i^{seq}$ represents the number of shared edges between pairs of faces. To encode this information, we introduce a set of learnable embeddings with a cardinality of $M_e+1$, where
\begin{equation}
    M_e := \max\left\{\max\left(\mathbf{FeF}_i\right) | 1\leq i \leq N\right\},
\end{equation}
which accounts for cases where faces have no shared edges.

\textbf{Encoder-decoder.}  We employ a Transformer-based VAE architecture \cite{vaswani2017attention, DBLP:journals/corr/KingmaW13} to model the distribution of edge-face sequences $\left\{ \mathbf{EF}_i^{seq} \right\}_{i=1}^N$. The encoder network $\mathcal{E}_\theta^{EF}$ processes the input sequence by combining three types of embeddings: positional embeddings \cite{vaswani2017attention}, shared-edges embeddings, and face ID embeddings. These combined embeddings are fed into the Transformer encoder to estimate the parameters of the latent distribution \cite{DBLP:journals/corr/KingmaW13}. The decoder network $\mathcal{D}_\theta^{EF}$ then maps this latent representation to output distributions over the discrete states $\left\{ 0, 1, \cdots, M_e \right\}$ at each sequence position, with the reconstructed sequence is sampled from these distributions. During inference, new edge-face sequences can be generated by sampling from the learned latent space and decoding through $\mathcal{D}_\theta^{EF}$.

\textbf{Loss function.} We employ the standard loss function used in VAE training, consisting of two components, 
\begin{equation} 
\begin{aligned} \mathcal{L}_{EF} = \frac{1}{N}\sum_{i=1}^N\Big( & CE\left( \mathbf{EF}_i^{seq}, \mathcal{D}_\theta^{EF}\left( \mathcal{E}_\theta^{EF}\left( \mathbf{EF}_i^{seq} \right) \right) \right)\\
& +D_{KL}\left( \mathcal{E}_\theta^{EF}\left( \mathbf{EF}_i^{seq} \right) || \mathcal{N}(0, \mathbf{I}) \right) \Big) 
\end{aligned} 
\end{equation} 
where $CE$ represents the cross-entropy loss between the input sequence and its reconstruction, and $D_{KL}$ represents the KL divergence between the learned latent distribution and the standard multivariate Gaussian prior $\mathcal{N}(0, \mathbf{I})$.

\subsection{Edge-vertex adjacency generation}
\label{sec: evGen}
After generating the edge-face adjacency, we know which edges are associated with each face, but the connectivity between these edges remains undetermined. The task of generating edge-vertex adjacency involves determining how these edges should be ordered and connected, while adhering to topological constraints ($C_1$-$C_3$). The $C_1$ constraint is inherently satisfied due to the properties of the $\left\{\mathbf{FeF}_i\right\}_{i=1}^{N}$ matrices. For the $C_2$ constraint, we enforce that the two endpoints of the same edge must not connect to each other. The most difficulty lies in the $C_3$ constraint, which requires that the edges of each face form closed loops. Since each edge is shared by two faces, forming loops for one face may compromise the ability to form valid loops for the adjacent face, making this a non-trivial challenge. To address this, we use a Transformer architecture \cite{vaswani2017attention} combined with a pointer network \cite{vinyals2015pointer} for edge-vertex adjacency generation.

\textbf{Sequential edge-vertex representation.} We reformulate edge connectivity generation (\ie, edge-vertex adjacencies) as a sequence generation task. Beginning from the face with the smallest ID, we serialize the edge connections for each face. For model $\mathcal{B}_i$, the sequence is defined as: 
\begin{equation} 
\label{eq: evSeq} 
\mathbf{EV}_i^{seq} := \left\{\mathbf{e}_{j_1}^i, \mathbf{e}_{j_2}^i, \dots, \mathbf{e}_{loop}, \mathbf{e}_{face}, \dots, \mathbf{e}_{face}\right\}, \end{equation} 
where $\mathbf{e}_{j_k}^i$ denotes the edge at index $j_k$ in $\mathcal{B}_i$, and $\mathbf{e}_{loop}$ and $\mathbf{e}_{face}$ are placeholders for the end of a loop and face, respectively. This sequence specifies the edge connectivity, \eg, $\mathbf{e}_{j_2}^i$ is connected to $\mathbf{e}_{j_1}^i$, and $\mathbf{e}_{j_3}^i$ is connected to $\mathbf{e}_{j_2}^i$. Each edge appears twice in $\mathbf{EV}_i^{seq}$ to account for its occurrence in both connected faces. To ensure sequence uniqueness, we assign a unique ID to each edge by lexicographically sorting the edges based on their connected face IDs. For edges that share the same faces, we randomly assign their order, mirroring the approach used for face ID assignment in \cref{sec: efGen}. After assigning edge IDs, we start with the smallest ID edge for each face and sequentially arrange the remaining connected edges. One further challenge is that each edge has two endpoints, meaning there are four possible configurations for how two edges might connect. To handle this, we duplicate each edge in our experiments, treating both endpoints individually. This ensures the edge connections is unambiguous. Further details on our serialization method are provided in the supplementary materials.

\textbf{Encoder.} The Transformer encoder $\mathcal{E}_\theta^{EV}$ processes the embeddings of edges in $\mathbf{E}_i$ along with two special tokens: $\mathbf{e}_{loop}$ and $\mathbf{e}_{face}$. Each edge embedding comprises three components: aggregated face features, endpoint embeddings, and shared-edge embeddings. For the first component, we use a Graph Convolutional Network (GCN) \cite{DBLP:conf/iclr/KipfW17} to extract features from the previously generated edge-face adjacencies, then average the features of each edge's two adjacent faces to form part of the edge embedding. For the endpoint embeddings, we introduce two distinct embeddings to represent each endpoint of the edge. Additionally, the shared-edges embeddings help to distinguish edges sharing the same faces as discussed in \cref{sec: efGen}. These three embeddings are summed to form the final edge embedding, which is then fed into the Transformer encoder, producing $2N_e^i + 2$ (accounting for edge duplication) contextual embeddings.  

\textbf{Pointer decoder.} Our decoder $\mathcal{D}_\theta^{EV}$ is designed to generate an edge-connection sequence based on the contextual embeddings from the encoder. We map each index in $\mathbf{EV}_i^{seq}$ to its corresponding contextual embedding, creating an aligned sequence that preserves the edge connection order. This sequence, enhanced with positional embeddings, is then processed by the Transformer decoder. Finally, a pointer network \cite{vinyals2015pointer, nash2020polygen} is applied to produce a probability distribution over possible connections, predicting the most likely edge connection at each position. 

\textbf{Loss function.} We employ cross-entropy loss between the predicted token probability distribution from the decoder and the ground truth sequence:
\begin{equation}
    \mathcal{L}_{EV}=\frac{1}{N}\sum_{i=1}^NCE\left( \mathbf{EV}_i^{seq}, \mathcal{D}_\theta^{EV}\left( \mathcal{E}_\theta^{EV}\left( \mathbf{E}_i \right), \mathbf{EV}_i^{seq} \right) \right),
\end{equation}

\textbf{Inference.} During inference, tokens are generated autoregressively with three key aspects: 1) Since each edge is shared by two faces, once the edge connections for the initial faces are determined, the connections for subsequent faces are partially fixed. Thus, we must identify these already-determined edges before processing each face. 2) When determining the next token, we first sample a candidate based on the predicted probability, then check it against topological constraints. If the candidate violates any constraints, it is discarded, and a new candidate is sampled from the remaining options. 3) Although topological constraints are not strictly enforced during training, experiments show high topological accuracy (see \cref{sec: uncond}), indicating the model effectively captures underlying structures. 

\subsection{Geometry generation}
\label{sec: geomGen} 

\begin{figure}
    \centering
    \includegraphics[width=\linewidth]{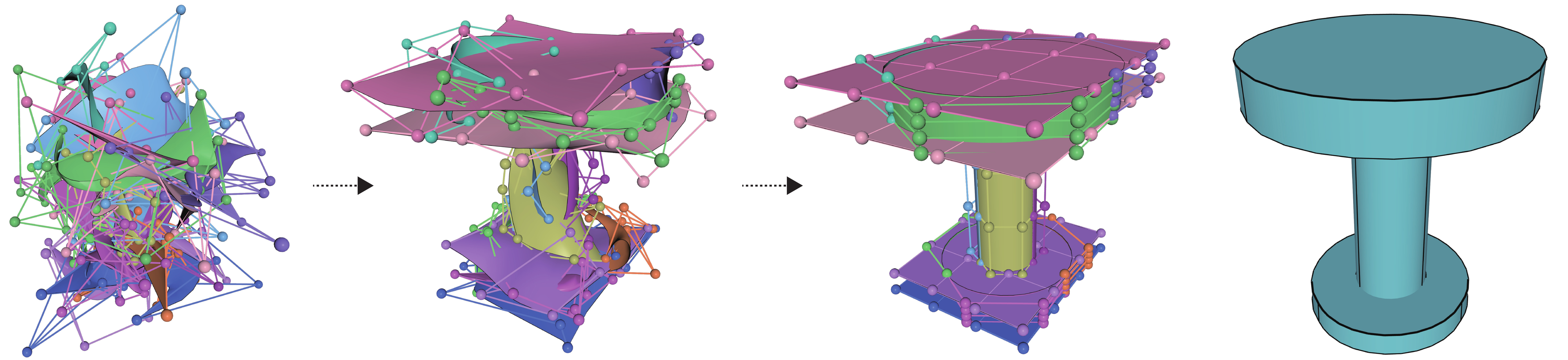}
    \caption{Illustration of B-spline control point generation via diffusion models. Our method learns the distribution of control points to establish face geometry in B-rep models, enabling precise geometric representation.} 
    \label{fig: ctrs} 
\end{figure}

As outlined in \cref{sec: overview}, our approach sequentially generates the face bounding boxes, vertex coordinates, edge geometries, and face geometries using Transformer-based diffusion models. We illustrate this process using face bounding boxes as an example. Let $\mathbf{FB}_i \in \mathbb{R}^{N_f^i \times 6}$ denote the bounding boxes representation for each face in $\mathcal{B}_i$. Following the standard diffusion process, we add noise to the $\left\{ \mathbf{FB}_i \right\}_{i=1}^N$ tokens and use a Transformer-based denoiser to predict this noise. A key feature of our approach is the integration of known topological structures. Specifically, when calculating attention scores between faces, we incorporate the number of shared edges (\ie, $\left\{ \mathbf{FeF}_i \right\}_{i=1}^N$), enhancing the model’s capacity to capture face-to-face interactions aligned with underlying topology \cite{ying2021transformers,vignac2023digress}. For vertex geometry generation $\mathbf{V}_i \in \mathbb{R}^{N_v^i \times 3}$, we include information on whether two vertices are connected by an edge, which informs the attention between vertices. Each edge is represented as a cubic B-spline curve with four control points, parameterized as $\mathbf{E}_i \in \mathbb{R}^{N_e^i \times 12}$ by these control points’ coordinates. For face geometry, we apply a similar method: each face is represented as a bi-cubic B-spline surface with a $4 \times 4$ grid of control points, resulting in face geometry $\mathbf{F}_i \in \mathbb{R}^{N_f^i \times 48}$ parameterized by the coordinates of these sixteen control points. Our diffusion models directly learns the distribution of these control points to capture precise geometric features. \cref{fig: ctrs} illustrates an example of the generation process for face control points.

\textbf{Post-processing.}  Since many surfaces correspond to basic types like planes or quadratic surfaces, when constructing the face geometry, we first attempt to fit the face’s boundary (discrete points sampled from the generated edge geometries) and interior (discrete points sampled from the generated face geometries) using basic primitives \cite{liu2024point2cad}. If the fitting error falls below a predefined threshold, we adopt the fitted primitives; otherwise, we resort to using the generated B-spline surface (\ie, generated face geometries). Finally, we use OpenCascade \cite{pyOCCT} functions to seamlessly sew the generated topology and geometries into a coherent B-rep solid.

\section{Experiments}
\subsection{Experiment setup}
\label{sec: experiment setup}
\textbf{Datasets.} We conduct experiments using the DeepCAD \cite{wu2021deepcad}, ABC \cite{koch2019abc}, and Furniture \cite{xu2024brepgen} datasets. B-reps with more than $50$ faces or faces with over $30$ edges are excluded. Following the filtering methods in \cite{wu2021deepcad, jayaraman2023solidgen, xu2024brepgen}, our final training dataset consists of $80,509$ DeepCAD B-reps, $198,522$ ABC B-reps, and $1,065$ Furniture B-reps.

\textbf{Network architecture.} For edge-face adjacency generation, we employ a $4$-layer Transformer encoder-decoder architecture with $128$-dimensional embeddings and $4$ attention heads. The edge-vertex adjacency model follows a similar architecture but uses $256$-dimensional embeddings. Both models utilize masked Transformer decoders and are trained using the teacher-forcing method \cite{williams1989learning}, where the ground-truth adjacency sequence is used as input to the decoder. During inference, tokens are generated in an autoregressive manner. For geometry generation, we employ four diffusion models, each featuring an $8$-layer Transformer-based denoiser with an embedding dimension of $512$ and $8$ attention heads. Additional details on architecture and training are in the supplementary materials.

\textbf{Evaluation metrics.} Following \cite{xu2024brepgen}, we use Distribution and CAD Metrics for a comprehensive evaluation. Distribution Metrics (Coverage (COV), Minimum Matching Distance (MMD), and Jensen-Shannon Divergence (JSD)) measure similarity between generated and ground-truth distributions, with COV and MMD are computed using Chamfer Distance (CD) in our experiment. CAD Metrics assess model quality and diversity, including Novel (models not in training set), Unique (models appearing only once), and Valid (watertight solid B-reps). Novel and Unique metrics are computed using the hashing procedure from \cite{jayaraman2023solidgen}, while our validity criteria require watertight, non-manifold structures with correct topology and geometry.

\subsection{Unconditional generation}
\label{sec: uncond}
\textbf{Topology generation evaluation.} We evaluate the performance of our topology generation model using three metrics: Novel, Unique, and Valid. These metrics follow CAD Metrics but are specifically applied to topology structure. To assess its performance, we generate 1,000 topologies using our topology generation network, which includes both edge-face and edge-vertex models. A topology is considered valid if it satisfies all the constraints ($C_1$, $C_2$ and $C_3$). Additionally, two topologies are considered equivalent if they share the same structural relationships among vertices, edges, and faces. As shown in \cref{tab: topology-metric}, DTGBrepGen achieves high novelty, uniqueness and validity rates across all the datasets, demonstrating its ability to generate diverse topological structures while maintaining validity and effectively capturing underlying patterns in B-rep topology generation.

\begin{table}
  \centering
  \begin{tabular}{@{}lccc@{}}
    \toprule
    Datasets & Novel $(\%)$ $\uparrow$ & Unique $(\%)$ $\uparrow$ & Valid $(\%)$ $\uparrow$\\
    \midrule
    DeepCAD & 85.01 & 82.27 & 92.10\\
    ABC  & 80.55 & 78.34 & 89.20 \\
    Furniture & 81.63 & 80.56 & 88.10\\
    \bottomrule
  \end{tabular}
  \caption{Quantitative evaluation of topology generation on DeepCAD, ABC, and Furniture datasets.}
  \label{tab: topology-metric}
\end{table}

\begin{figure*}
    \centering
    \begin{subfigure}{0.32\textwidth}
        \centering
        \includegraphics[width=\linewidth]{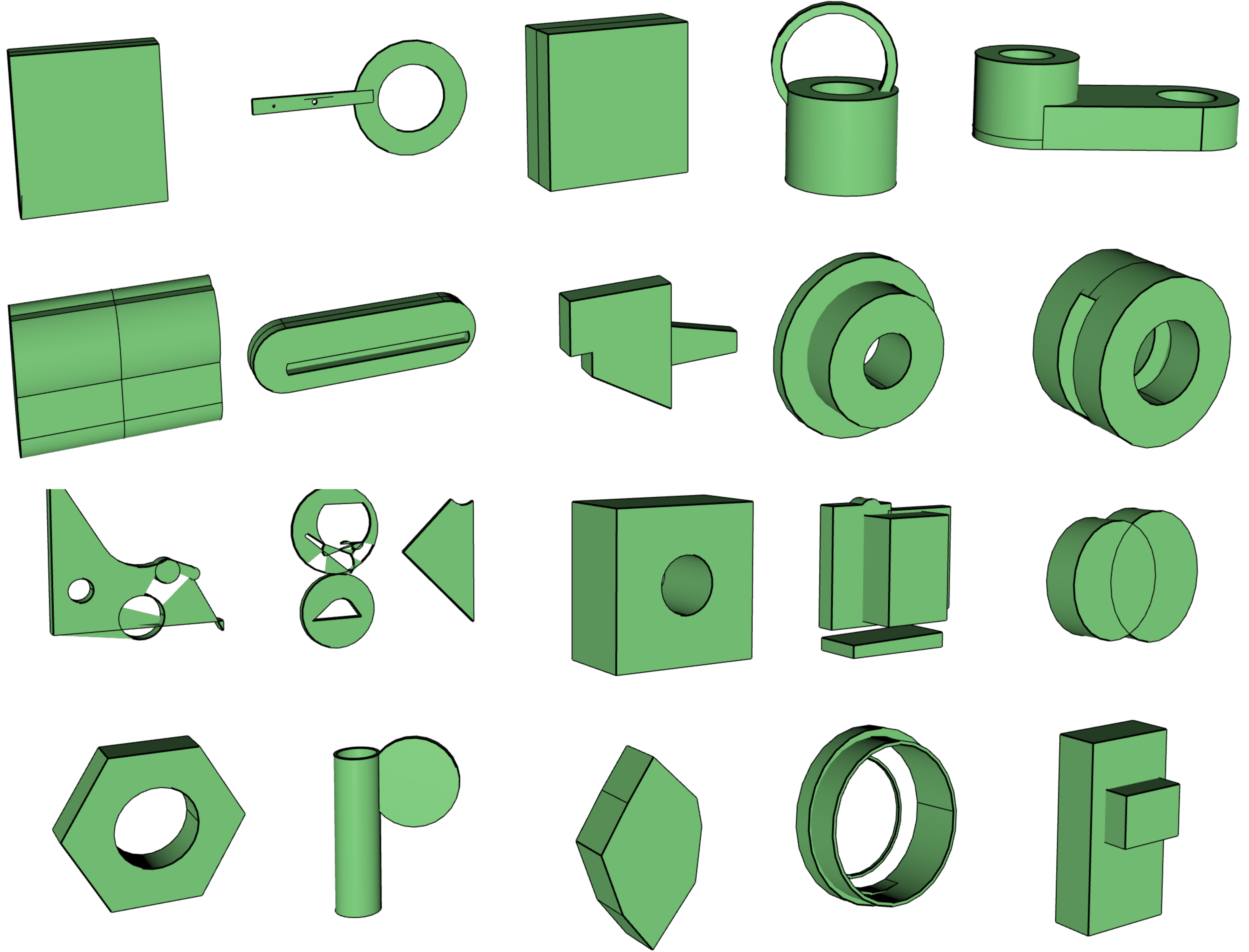}
        \caption{DeepCAD}
    \end{subfigure}
    \vrule width 0.5pt
    \begin{subfigure}{0.32\textwidth}
        \centering
        \includegraphics[width=\linewidth]{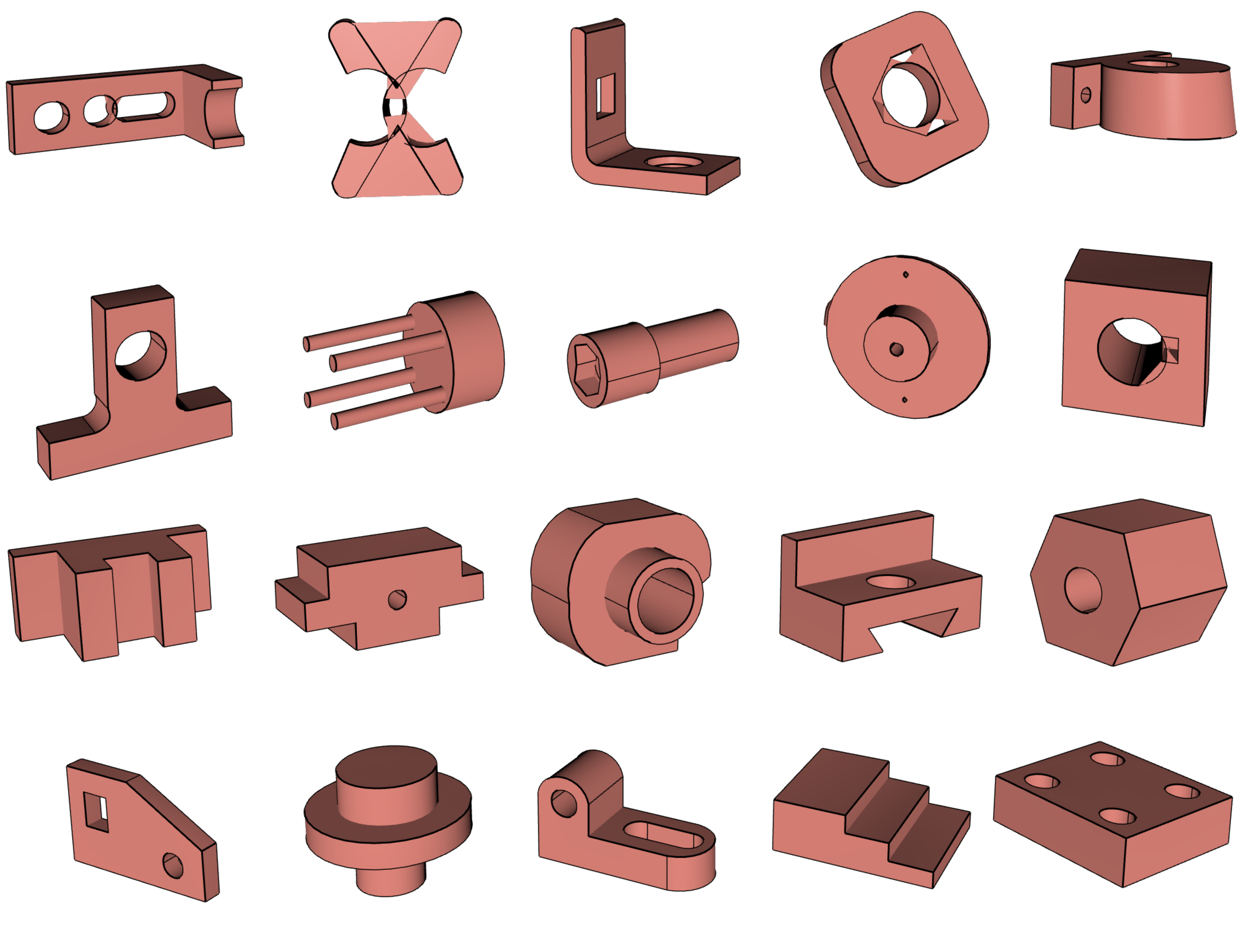}
        \caption{BrepGen}
    \end{subfigure}
    \vrule width 0.5pt
    \begin{subfigure}{0.32\textwidth}
        \centering
        \includegraphics[width=\linewidth]{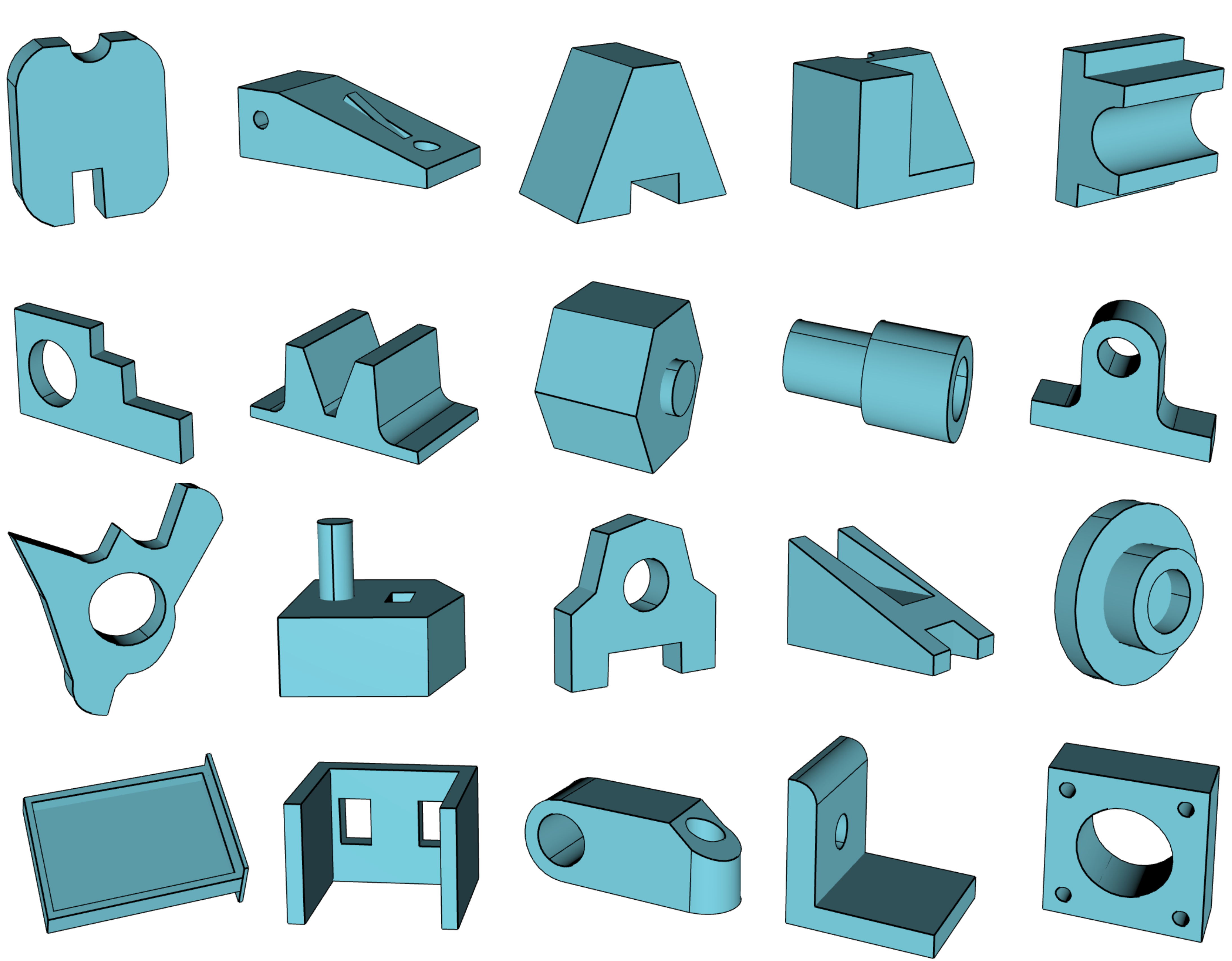}
        \caption{DTGBrepGen (ours)}
    \end{subfigure}
    \caption{Qualitative comparison of B-rep models generated by our method, DeepCAD \cite{wu2021deepcad}, and BrepGen \cite{xu2024brepgen} on the DeepCAD dataset.}
    \label{fig: comparison}
\end{figure*}

\textbf{Comparison with baseline methods.} We compare our approach with two representative B-rep generation methods: DeepCAD \cite{wu2021deepcad} and the state-of-the-art BrepGen \cite{xu2024brepgen}. For DeepCAD, we evaluate B-rep models reconstructed from generated sketches and extrusion sequences. We randomly sample $3,000$ generated and $1,000$ reference B-rep models. For each model, we sample $2,000$ points from its surface to compute the Distribution Metrics. The CAD Metrics are calculated using the generated $3,000$ B-rep models. As shown in \cref{tab: comparison}, DTGBrepGen achieves better performance across almost all the metrics, with particularly substantial improvements in the Valid metric, demonstrating our method's superior capability in generating well-formed and watertight B-rep models. The qualitative comparison in \cref{fig: comparison} further illustrates our DTGBrepGen's ability to generate more realistic and geometrically precise B-rep models compared to baseline approaches. 

\begin{table}
  \centering
  \begingroup
   \setlength{\tabcolsep}{2pt} 
  \begin{tabular}{@{}lcccccc@{}}
    \toprule
    Method & \multicolumn{6}{c}{DeepCAD} \\
    & COV$\uparrow$ & MMD$\downarrow$ & JSD$\downarrow$ & Novel$\uparrow$ & Unique$\uparrow$ & Valid$\uparrow$ \\
    \midrule
    DeepCAD & 70.81 & 1.31 & 1.79 & 93.80 & 89.79 & 58.10 \\
    BrepGen & 72.38 & 1.13 & 1.29 & 99.72 & \textbf{99.18} & 68.23 \\
    Ours & \textbf{74.52} & \textbf{1.07} & \textbf{1.02} & \textbf{99.79} & 98.94 & \textbf{79.80} \\
    Ours$^*$ & 72.86 & 1.18 & 1.09 & 99.59 & 99.02 & 71.77 \\
    \midrule
    Method & \multicolumn{6}{c}{ABC} \\
    & COV$\uparrow$ & MMD$\downarrow$ & JSD$\downarrow$ & Novel$\uparrow$ & Unique$\uparrow$ & Valid$\uparrow$ \\
    \midrule
    BrepGen & 64.29 & 1.53 & 1.86 & 99.68 & 99.05 & 47.11 \\
    Ours & \textbf{71.13} & \textbf{1.33} & \textbf{1.59} & \textbf{99.73} & \textbf{99.12} & \textbf{62.08} \\
    Ours$^*$ & 69.60 & 1.37 & 1.65 & 99.56 & 98.97 & 56.63 \\
    \bottomrule
  \end{tabular}
  \endgroup
  \caption{Quantitative comparison with baseline methods on DeepCAD and ABC datasets using Distribution Metrics (COV, MMD, JSD) and CAD Metrics (Novel, Unique, Valid). Note that MMD and JSD values are multiplied by $10^2$, while COV, Novel, Unique, and Valid are expressed as percentages. Ours$^*$ corresponds to an ablation variant of DTGBrepGen where B-spline representations are replaced with point cloud-based latent encoding. }
  \label{tab: comparison}
\end{table}

\subsection{Conditional generation}
Our framework supports conditional generation by incorporating additional contextual information. We demonstrate this capability through two distinct tasks: class-conditioned generation and point cloud-conditioned generation.

\begin{figure*}
    \centering
    \begin{tikzpicture}
        \node[anchor=north west] (img) at (0,0) {\includegraphics[width=\linewidth]{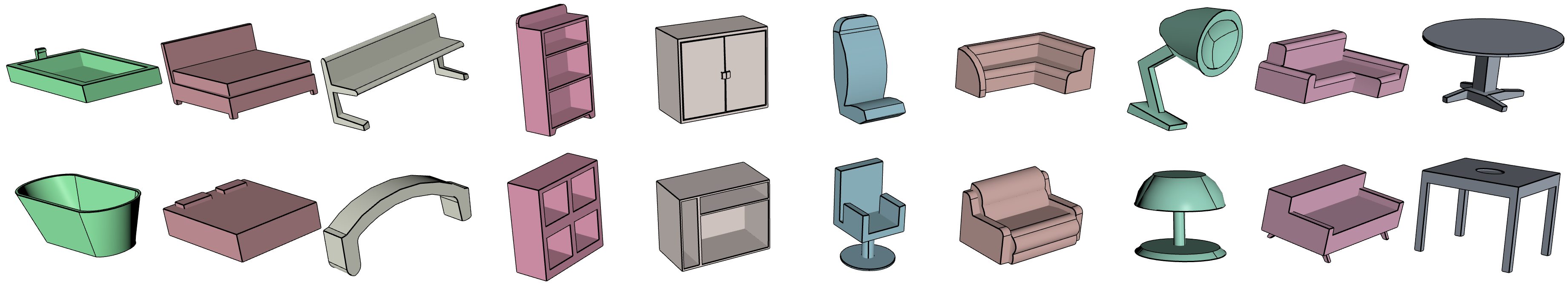}};
        \node[anchor=north west, xshift=0.6cm, yshift=-3.2cm] {\footnotesize bathtub};
        \node[anchor=north west, xshift=2.5cm, yshift=-3.2cm] {\footnotesize bed};
        \node[anchor=north west, xshift=4.2cm, yshift=-3.2cm] {\footnotesize bench};
        \node[anchor=north west, xshift=5.7cm, yshift=-3.2cm] {\footnotesize bookshelf};
        \node[anchor=north west, xshift=7.7cm, yshift=-3.2cm] {\footnotesize cabinet};
        \node[anchor=north west, xshift=9.5cm, yshift=-3.2cm] {\footnotesize chair};
        \node[anchor=north west, xshift=11.2cm, yshift=-3.2cm] {\footnotesize couch};
        \node[anchor=north west, xshift=12.9cm, yshift=-3.2cm] {\footnotesize lamp};
        \node[anchor=north west, xshift=14.6cm, yshift=-3.2cm] {\footnotesize sofa};
        \node[anchor=north west, xshift=16.3cm, yshift=-3.2cm] {\footnotesize table};
    \end{tikzpicture}
    \caption{Examples of class-conditioned generation results for different furniture categories. Each column shows two instances from the same category, demonstrating our method's ability to capture category-specific features while ensuring structural diversity.}
    \label{fig: class}
\end{figure*}
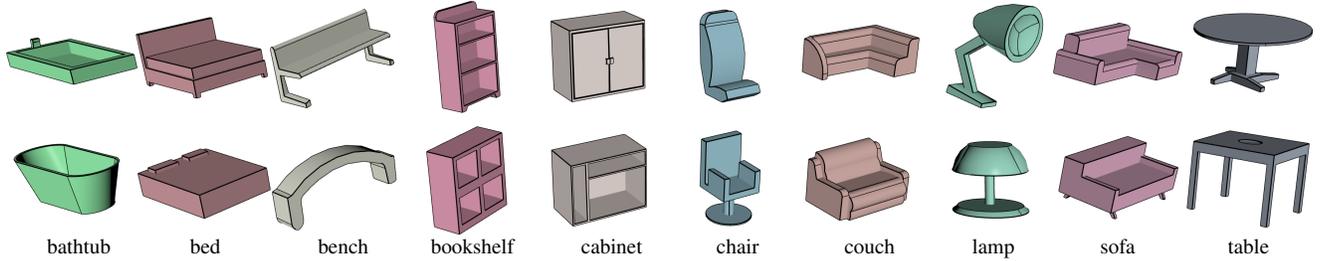

\textbf{Class-conditioned generation.} We evaluate class-conditioned generation on the Furniture B-rep dataset ($10$ categories), by assigning each category a learnable embedding added to the input tokens and using classifier-free guidance \cite{ho2022classifier} for controlled generation. As shown in \cref{fig: class}, our method effectively captures category-specific features while maintaining structural diversity. For quantitative assessment, we compare our approach to BrepGen. Given data limitations in the Furniture dataset, we focus Distribution Metrics on the four largest categories (table, chair, bed, sofa). As seen in \cref{tab: furniture}, our method consistently outperforms BrepGen across all the metrics in these categories. We also evaluate generation quality with CAD Metrics, reporting only the Valid metric due to consistently high Novel and Unique scores ($98-100\%$). In \cref{tab: furniture-valid}, our approach achieves approximately $10\%$ higher Valid scores across all categories compared to BrepGen. Both methods show lower validity rates in certain categories, such as Lamp, likely due to limited training samples. Overall, these results emphasize DTGBrepGen's effectiveness in generating structurally valid, category-specific furniture models.

\begin{table}[htbp]
  \centering
  \begingroup
  \setlength{\tabcolsep}{4pt}
  \begin{tabular}{@{}lccc|ccc@{}}
    \toprule
    Method & \multicolumn{3}{c|}{Table} & \multicolumn{3}{c}{Chair} \\
    & COV$\uparrow$ & MMD$\downarrow$ & JSD$\downarrow$ & COV$\uparrow$ & MMD$\downarrow$ & JSD$\downarrow$ \\
    \midrule
    BrepGen & 65.08 & 0.97 & 4.30 & 61.24 & 0.91 & 3.50 \\
    Ours & \textbf{70.20} & \textbf{0.63} & \textbf{2.69} & \textbf{69.48} & \textbf{0.67} & \textbf{2.36} \\
    \midrule
    Method & \multicolumn{3}{c|}{Bed} & \multicolumn{3}{c}{Sofa} \\
    & COV$\uparrow$ & MMD$\downarrow$ & JSD$\downarrow$ & COV$\uparrow$ & MMD$\downarrow$ & JSD$\downarrow$ \\
    \midrule
    BrepGen & 59.30 & 0.75 & 3.66 & 67.82 & 0.45 & 1.86 \\
    Ours & \textbf{71.11} & \textbf{0.58} & \textbf{2.47} & \textbf{73.46} & \textbf{0.42} & \textbf{1.75} \\
    \bottomrule
  \end{tabular}
  \endgroup
  \caption{Comparison of Distribution Metrics across the four most populous categories in the Furniture dataset for class-conditioned generation. Note that MMD and JSD values are multiplied by $10^2$.}
  \label{tab: furniture}
  
  \vspace{1.5em}  
  
  \begingroup
  \setlength{\tabcolsep}{4pt}
  \begin{tabular}{@{}lccccc@{}}
    \toprule
    Method & Bathtub & Bed & Bench & Bookshelf & Cabinet \\
    \midrule
    BrepGen & 28.50 & 57.63 & 66.03 & 33.53 & 34.57 \\
    Ours & \textbf{49.82} & \textbf{67.87} & \textbf{69.44} & \textbf{50.82} & \textbf{69.16} \\
    \midrule
    Method & Chair & Couch & Lamp & Sofa & Table \\
    \midrule
    BrepGen & 72.45 & 70.54 & 29.53 & 76.25 & 55.67 \\
    Ours & \textbf{80.17} & \textbf{73.45} & \textbf{36.62} & \textbf{79.93} & \textbf{68.27} \\
    \bottomrule
  \end{tabular}
  \endgroup
  \caption{Comparison of Valid metric ($\%$) between our method and BrepGen across all categories in the Furniture dataset.}
  \label{tab: furniture-valid}
\end{table}

\textbf{Point cloud-conditioned generation.} We adapt our Transformer for point cloud-conditioned generation by integrating a PointNet++ \cite{qi2017pointnet++} that encodes $2,000$ sampled points into a $512$-dimensional embedding. Instead of using cross-attention, this embedding is added directly to each token before being input to the Transformer, allowing the network to integrate point cloud information without altering the core Transformer architecture. As shown in \cref{fig: point2brep}, our model generates diverse yet coherent B-reps that faithfully preserve the input geometry's characteristics. These results validate the effectiveness of our approach in generating meaningful variations while preserving the essential features of the conditioning data. While effective in most cases, we observe that for point clouds with highly intricate geometric details, generation quality can be affected, indicating potential for future work on improving the encoding of fine-grained geometric features in complex point clouds.

\begin{figure}
    \centering
    \includegraphics[width=\linewidth]{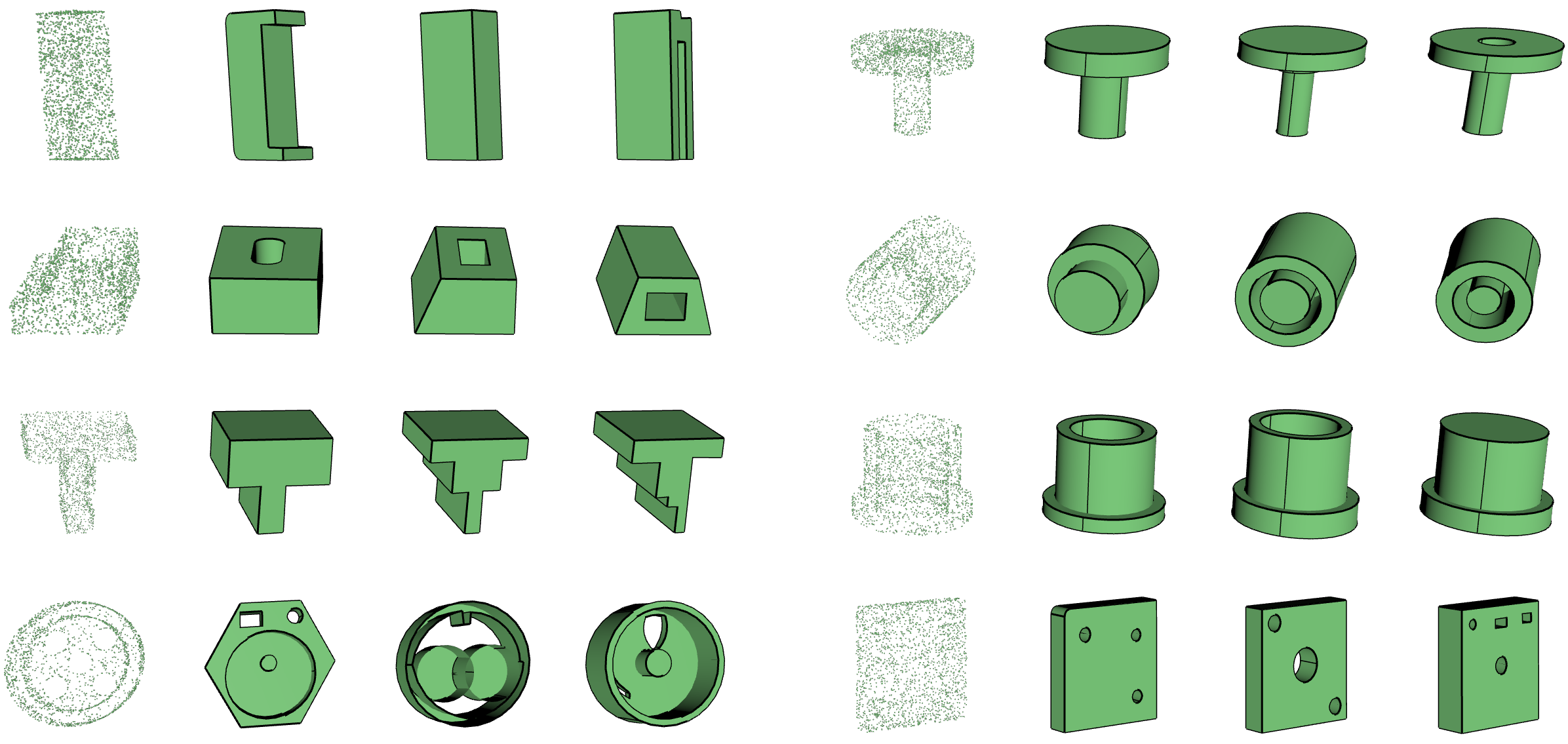}
    \caption{B-rep models generated based on input point clouds. Each example consists of a point cloud alongside three corresponding generated B-reps.} 
    \label{fig: point2brep} 
\end{figure}

\subsection{Discussion}
\textbf{Ablation study.} To assess the impact of using B-spline parametrization for edges and faces, we conduct an ablation study by replacing the B-spline geometric representation with discrete points \cite{jayaraman2021uv, jayaraman2023solidgen, xu2024brepgen}. Specifically, within the context of edge and face generation, we substitute the B-spline representation with a latent representation obtained from a VAE-based point cloud encoder. As shown in \cref{tab: comparison}, our original approach, which employs B-spline representations, outperforms the discrete point-based variant across all the metrics. These results underscore the effectiveness of directly learning control point distributions, which enables our model to capture the underlying geometric structure more accurately and effectively. Qualitative comparison results are available in the supplementary materials.

\begin{figure}
    \centering
    \begin{subfigure}{0.48\linewidth}
        \centering
        \includegraphics[width=\linewidth]{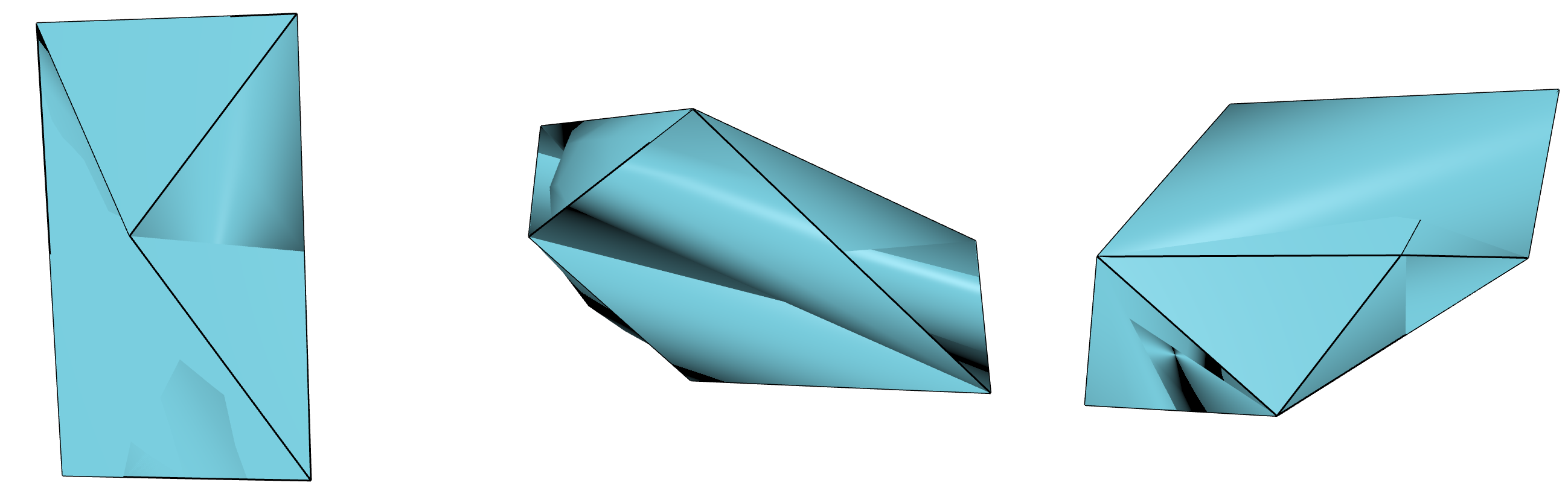}
        \caption{}
        \label{fig: bug}
    \end{subfigure}%
    \hspace{0.01\linewidth}
    \vrule width 0.5pt 
    \begin{subfigure}{0.48\linewidth}
        \centering
        \includegraphics[width=\linewidth]{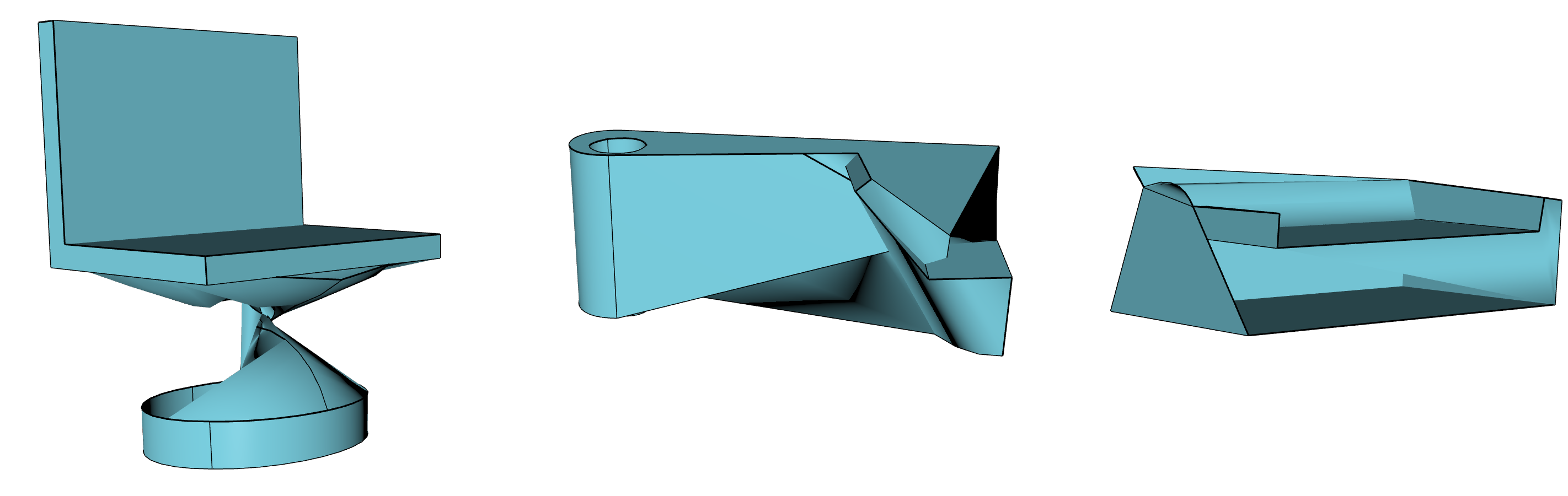}
        \caption{}
        \label{fig: fail}
    \end{subfigure}
\caption{Failure cases. (a) A valid topology sampled from our topology network that leads to unsuccessful geometry generation across multiple attempts. (b) Instances where our method produces non-watertight solids and self-intersecting geometries.}
    \label{fig:point2brep}
\end{figure}

\textbf{Limitations and future work.} While DTGBrepGen demonstrates significant improvements over existing methods, several limitations remain to be addressed. First, although we achieve higher B-rep validity rates, the overall validity still leaves room for improvement. This limitation mainly arises from a gap between the learned and true topology distributions. As shown in \cref{fig: bug}, geometry generation occasionally fails even with valid sampled topologies. This phenomenon indicates that while our topology network can reconstruct topological structures with high accuracy (see \cref{tab: topology-metric}), not all valid topologies are necessarily suitable for geometry generation. This observation is further supported by our experiments using ground-truth topologies, which yield a higher B-rep validity rate (see \cref{tab: gt-topology}). Additionally, similar to existing methods, DTGBrepGen faces challenges when generating complex models, particularly those with intricate geometric details and complicated structural relationships. Future work could focus on advanced topology validation mechanisms, improved designs to bridge topology and geometry generation, and hierarchical strategies to better handle complex models. 

\begin{table}
  \centering
  \begin{tabular}{@{}lccc@{}}
    \toprule
    Datasets & DeepCAD & ABC & Furniture \\
    \midrule
     Valid $(\%) \uparrow$ & 90.24 & 84.61 & 82.59 \\
    \bottomrule
  \end{tabular}
  \caption{B-rep validity rates for geometry generation using ground-truth topologies.}
  \label{tab: gt-topology}
\end{table}

\section{Conclusion}
In this paper, we presented DTGBrepGen, a framework for B-rep generation that decouples topology and geometry, addressing key limitations in existing methods. By separately modeling topological and geometric attributes, DTGBrepGen enhances the validity and diversity of generated CAD models, offering a promising tool for CAD design exploration and synthesis. We believe this decoupled methodology contributes a valuable perspective to the field, potentially supporting future advances in automatic B-rep generation.

{
    \small
    \bibliographystyle{ieeenat_fullname}
    \bibliography{main}
}

\clearpage
\appendix
\section*{Appendix}
\section{Method and network architectures}
\subsection{Sequential edge-vertex representation}
In this section, we elaborate on the details of our method for serializing the edge-vertex adjacency relationships in the B-rep dataset during the training phase. Following the notations introduced in the main paper, we present our approach using the B-rep model $\mathcal{B}_i$ as an example. The B-rep model $\mathcal{B}_i$ consists of $N_f^i$ faces, indexed as $0, 1, \dots, N_f^i-1$. Each face contains a set of edges, denoted as $\mathbf{FE}_i$, which is the dual representation of the edge-face relationship $\mathbf{EF}_i$. To ensure unambiguous edge-to-edge connectivity, we introduce a unique indexing scheme for edge endpoints. Specifically, for the $j$-th edge $\mathbf{e}_j^i$, we assign IDs $2j$ and $2j+1$ to its two endpoints, referred to as even and odd endpoints respectively. In the main paper, we refer to this process as edge duplication, which we will leverage in \cref{sec: ev}. Consequently, the B-rep model $\mathcal{B}_i$ contains $2N_e^i$ unique endpoint IDs.To handle the many-to-one correspondence between endpoints and vertices, we define a mapping vector $\Tilde{\mathbf{V}}_i$ of length $2N_e^i$, where $\Tilde{\mathbf{V}}_i[j]$ stores the vertex ID in $\mathbf{V}_i$ corresponding to the $j$-th endpoint. The mapping vector is initialized with $-1$ values and updated progressively during computation. 

\begin{figure}
	\centering
	\includegraphics[width=0.3\linewidth]{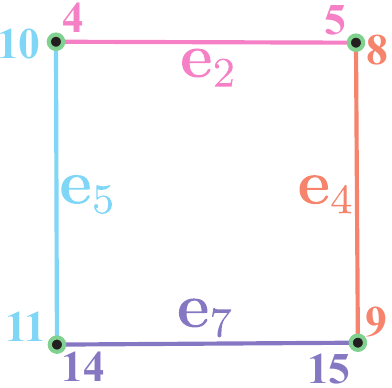}
	\caption{Illustration of our sequential edge-vertex representation. A simple loop consisting of four edges ($\mathbf{e}_2$, $\mathbf{e}_4$, $\mathbf{e}_7$, and $\mathbf{e}_5$) is shown with their endpoint IDs labeled. The loop is serialized into sequence $(4,8,15,11,-1)$.} 
	\label{fig: seq} 
\end{figure}

The serialization process starts by initializing an empty sequence $\mathbf{EV}_i^{seq}$. For each face $\mathbf{f}_k^i$ beginning with $\mathbf{f}_0^i$, we first identify the edge with the minimum ID as our starting point and add its even endpoint ID to $\mathbf{EV}_i^{seq}$. Following the loop structure, we iteratively process connected edges by selecting the connecting edge with the smaller ID at each step and adding its corresponding endpoint ID to $\mathbf{EV}_i^{seq}$. When a loop is completed, we append a loop-end flag $\mathbf{e}_{loop}$ (numerically represented as -1). This process repeats for any remaining edges within the face, treating each as a new starting point for subsequent loops. After processing all edges in a face, we append a face-end flag $\mathbf{e}_{face}$ (numerically represented as -2) before moving to the next face. \cref{fig: seq} illustrates our approach with a simple example: a loop composed of edges $\mathbf{e}_2$, $\mathbf{e}_4$, $\mathbf{e}_7$, and $\mathbf{e}_5$. The endpoint IDs are labeled adjacent to their corresponding vertices (\eg, endpoint IDs $4$ and $10$ map to the same vertex). Following our serialization procedure, this loop generates the sequence $(4, 8, 15, 11, -1)$. The complete algorithm is formally presented in \cref{alg: topo_seq}.

\begin{algorithm}
	\caption{Sequential Edge-Vertex Representation}
	\label{alg: topo_seq}
	\begin{algorithmic}[1]
		\Require Face-Edge adjacency matrix $\mathbf{FE}_i$, Edge-Vertex adjacency matrix $\mathbf{EV}_i$
		\Ensure Edge-Vertex sequence $\mathbf{EV}_i^{seq}$
		\Function{opp\_vert}{$\mathbf{v}, \mathbf{e}$}
		\Comment{Returns the vertex ID opposite to $\mathbf{v}$ in edge $\mathbf{e}$}
		\EndFunction
		\Function{opp\_endpnt}{$\mathbf{ep}$}
		\Comment{Returns the endpoint ID opposite to $\mathbf{ep}$}
		\EndFunction
		
		\State Initialize number of faces $N_f^i$ and number of edges $N_e^i$
		\State Initialize $\mathbf{EV}_i^{seq} \gets \emptyset$ 
		\State Initialize $\mathbf{e}_{loop} \gets -1$, $\mathbf{e}_{face} \gets -2$ 
		\State Initialize $\Tilde{\mathbf{V}}_i \gets \left\{-1\right\}^{2N_e^i}$ \Comment{Mapping vector}
		
		\For{face index $j \gets 0$ to $N_f^i-1$}
		\State $\mathbf{E}_{rest} \gets \mathbf{FE}_i[j]$ \Comment{Edges of the current face}
		
		\While{$\mathbf{E}_{rest} \neq \emptyset$}
		\State $\mathbf{e}_{cur} \gets \min(\mathbf{E}_{rest})$ \Comment{Starting edge}
		\State Remove $\mathbf{e}_{cur}$ from $\mathbf{E}_{rest}$
		\State $\mathbf{ep} \gets 2\mathbf{e}_{cur}$ \Comment{Current endpoint}
		\State Append $\mathbf{ep}$ to $\mathbf{EV}_i^{seq}$
		\State $\mathbf{v}_{cur} \gets$ corresponding vertex of $\mathbf{ep}$
		\State $\mathbf{v}_{opp} \gets \texttt{opp\_vert}(\mathbf{v}_{cur}, \mathbf{e}_{cur})$ 
		\State $\Tilde{\mathbf{V}}_i[\mathbf{ep}] \gets \mathbf{v}_{cur}$
		\State $\Tilde{\mathbf{V}}_i[\texttt{opp\_endpnt}(\mathbf{ep})] \gets \mathbf{v}_{opp}$
		\State $\mathbf{v}_{cur} \gets \mathbf{v}_{opp}$
		\While{True} \Comment{Process current loop}
		\State Find $\mathbf{e}_{next} \in \mathbf{E}_{rest}$ connected to $\mathbf{v}_{cur}$
		\If{$\mathbf{e}_{next} = \mathbf{e}_{start}$} \Comment{Loop closed}
		\State Append $\mathbf{e}_{loop}$ to $\mathbf{EV}_i^{seq}$
		\State \textbf{break}
		\Else
		\State $\mathbf{ep} \gets$ corresponding endpoint to $\mathbf{v}_{cur}$ 
		\State Append $\mathbf{ep}$ to $\mathbf{EV}_i^{seq}$
		\State $\mathbf{v}_{opp} \gets \texttt{opp\_vert}(\mathbf{v}_{cur}, \mathbf{e}_{next})$
		\State $\Tilde{\mathbf{V}}_i[\mathbf{ep}] \gets \mathbf{v}_{cur}$
		\State $\Tilde{\mathbf{V}}_i[\texttt{opp\_endpnt}(\mathbf{ep})] \gets \mathbf{v}_{opp}$
		\State Remove $\mathbf{e}_{next}$ from $\mathbf{E}_{rest}$
		\State $\mathbf{e}_{cur} \gets \mathbf{e}_{next}$
		\State $\mathbf{v}_{cur} \gets \mathbf{v}_{opp}$
		\EndIf
		\EndWhile
		\EndWhile
		\State Append $\mathbf{e}_{face}$ to $\mathbf{EV}_i^{seq}$ \Comment{Mark end of the face}
		\EndFor
		
		\Return $\mathbf{EV}_i^{seq}$
		
	\end{algorithmic}
\end{algorithm}

\subsection{Details of edge-face adjacency generation}
\label{sec: ef}

\begin{figure}
	\centering
	\includegraphics[width=\linewidth]{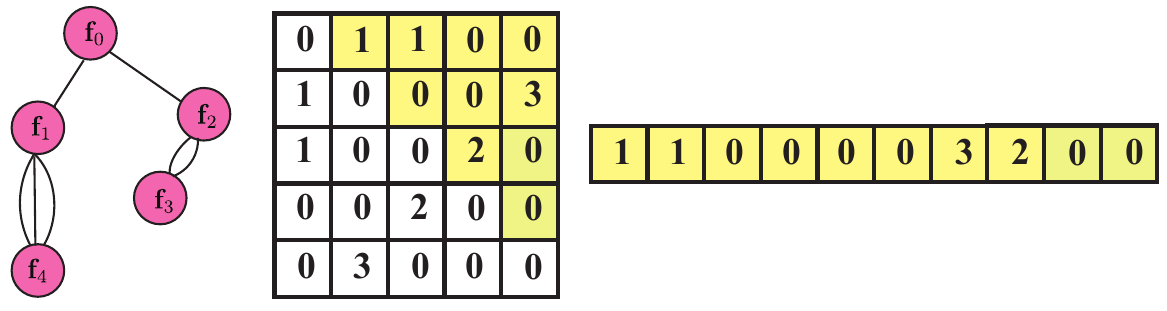}
	\caption{An illustrative example of three topologically equivalent representations of the edge-face relationships in a B-rep: (1) an edge-face graph, where nodes represent faces and connections represent shared edges; (2) a face-edge-face matrix, where entries indicate the number of shared edges between two faces; and (3) a serialized edge-face sequence, obtained by flattening the upper triangular part of the matrix in row-major order.} 
	\label{fig: graph} 
\end{figure}

\textbf{Graph-sequence representation.} \cref{fig: graph} provides a simple example illustrating three topologically equivalent representations of edge-face relationships in a B-rep: (1) the edge-face graph, (2) the face-edge-face matrix, and (3) the edge-face sequence. The example is based on a B-rep with five faces, where pairs of faces share zero, one, or multiple edges. As described in the main paper, the face-edge-face matrix encodes the number of shared edges between any two faces in the model. A matrix entry of $0$ indicates no shared edges between the corresponding pair of faces. Exploiting the symmetry of this matrix, we restrict our attention to its upper triangular part (excluding the diagonal) for further processing. The upper triangular elements are then serialized into a sequence by flattening them in row-major order, ensuring that the sequence retains the complete topological adjacency information of the faces. Our Edge-Face Adjacency generation model $\mathcal{E}_{\theta}^{EF}$ is designed to learn the distribution of these sequences, effectively capturing the underlying topological patterns present in the edge-face relationships of B-rep models. This sequential representation simplifies the learning task while retaining all necessary adjacency information.

\textbf{Embedding.} To standardize the varying lengths of sequences $\left\{\mathbf{EF}_i^{seq}\right\}_{i=1}^N$, we pad each face-edge-face matrix $\mathbf{FeF}_i$ to a fixed size $M_f \times M_f$ using zeros, where $M_f$ is the maximum number of faces across $\left\{\mathcal{B}_i\right\}_{i=1}^N$. The edge-face sequence is then extracted from the upper triangular part of the padded matrix in a row-major order. For simplicity, we continue to denote the padded matrix and the extracted sequence as $\mathbf{FeF}_i$ and $\mathbf{EF}_i^{seq}$, respectively. Note that the length of the serialized edge-face sequence $\mathbf{EF}_i^{seq}$ is $\frac{M_f(M_f-1)}{2}$. We define the embedding function in the edge-face model as:
\begin{equation} 
	EM_\theta^{EF}: \{0, 1, \ldots, M_e\}^{\frac{M_f(M_f-1)}{2}} \rightarrow \mathbb{R}^{\frac{M_f(M_f-1)}{2} \times d_{ef}},
\end{equation}
where this function maps each integer in $\mathbf{EF}_i^{seq}$ to a $d_{ef}$-dimensional embedding, capturing various semantic and structural properties. Let $SE_\theta$, $POS_\theta$, and $FID_\theta$ represent the shared-edges embedding, positional embedding, and face ID embedding, respectively. The final embedding is computed as:
\begin{equation} 
	\begin{aligned}
		EM_\theta^{EF}(\mathbf{EF}_i^{seq}) &= SE_\theta(\mathbf{EF}_i^{seq}) + POS_\theta(\mathbf{EF}_i^{seq}) \\
		&\quad + FID_\theta(\mathbf{EF}_i^{seq}), 
	\end{aligned}
	\label{eq: ef-embed} 
\end{equation} 
where the shared-edges embedding is defined as:
\begin{equation}
	SE_\theta(\mathbf{EF}_i^{seq}) = (W_\theta \cdot \text{onehot}(\mathbf{EF}_i^{seq}))^T,
\end{equation}
Here, $W_\theta \in \mathbb{R}^{d_{ef}\times (M_e+1)}$ is a learnable matrix and $\text{onehot}(\mathbf{EF}_i^{seq}) \in \{0, 1\}^{(M_e+1) \times \frac{M_f(M_f-1)}{2}}$ represents the one-hot encoding of $\mathbf{EF}_i^{seq}$. The positional embedding $POS_\theta$ indicates the position index of each token in the sequence, implemented using the sinusoidal positional encoding scheme proposed in \cite{vaswani2017attention}. For the face ID embedding, we assign each face a learnable $d_{ef}$-dimensional embedding vector. The embedding of an edge is computed as the average of the embeddings of the two faces it connects. Formally, for the $k$-th element in the sequence $\mathbf{EF}_i^{seq}$, the corresponding face ID embedding $FID_\theta(\mathbf{EF}_i^{seq})[k]$ is defined as:
\begin{equation}
	\begin{aligned}
		FID_\theta(\mathbf{EF}_i^{seq})[k] = &\frac{(V_\theta \cdot \text{onehot}(k_{row}))^T}{2} + \\
		&\frac{(V_\theta \cdot \text{onehot}(k_{col}))^T}{2},
	\end{aligned}
\end{equation}
where $V_\theta \in \mathbb{R}^{d_{ef} \times M_f}$ is a learnable matrix, and $k_{row}$, $k_{col}$ represent the row and column indices of the $k$-th element in $\mathbf{FeF}_i$'s upper triangular portion. 

\textbf{Training workflow.} During training, the sequence $\mathbf{EF}_i^{seq}$ is fed into the Transformer encoder, which outputs a contextual embedding with shape $\frac{M_f(M_f - 1)}{2} \times d_{ef}$. To obtain the latent code, we compute the mean of the embeddings for all tokens in the encoder's output. This aggregation step ensures that the latent code encapsulates global contextual information from the input sequence. In the decoding phase, a special token is prepended to the sequence $\mathbf{EF}_i^{seq}$ as a ``begin'' token, while the last token of the sequence is removed. The modified sequence is then passed through the embedding layer, as discussed in the previous paragraph, to obtain its embeddings. These embeddings are combined with the latent code and fed into the masked Transformer decoder, which outputs a probability distribution over all possible tokens for each position in the sequence. This architecture aligns with the framework of a standard Variational Autoencoder (VAE) \cite{DBLP:journals/corr/KingmaW13}, where the encoder and decoder are jointly optimized. \cref{tab: ef-shape} summarizes the output shapes of each module during the training phase.

\textbf{Autoregressive generation.} During inference, we sample a latent code from the learned latent space and initialize the sequence with a ``begin'' token. The initialized sequence, combined with the latent code, is fed into the Transformer decoder to obtain the probability distribution of the first token. We sample from this distribution to generate the first token of the sequence. Subsequently, this sampled token is appended to the current sequence, which is then passed through the decoder to predict the distribution of the next token. This process is repeated iteratively until the sequence reaches the predefined length (\ie, $\frac{M_f(M_f-1)}{2}$).

\begin{table*}
	\centering
	\begin{tabular}{ccccc}
		\toprule
		& Input & Embedding & (Masked) Transformer blocks & Output layer \\ 
		\midrule
		Encoder & $\frac{M_f(M_f-1)}{2}$  & $\frac{M_f(M_f-1)}{2} \times d_{ef}$ & $\frac{M_f(M_f-1)}{2} \times d_{ef}$   &  $d_{ef}$  \\ 
		Decoder & $\frac{M_f(M_f-1)}{2}, d_{ef}$ & $\frac{M_f(M_f-1)}{2} \times d_{ef}, d_{ef}$  & $\frac{M_f(M_f-1)}{2} \times d_{ef}$  & $\frac{M_f(M_f-1)}{2} \times (M_e+1)$  \\ 
		\bottomrule
	\end{tabular}
	\caption{Summary of output shapes for each module of our edge-face model during the training phase.}
	\label{tab: ef-shape}
\end{table*}

\subsection{Details of edge-vertex adjacency generation}
\label{sec: ev}
\textbf{Face feature extraction.} To integrate the information from the edge-face adjacency (edge-face graph) generated in the previous step, we employ a graph convolutional network (GCN) \cite{DBLP:conf/iclr/KipfW17}. Specifically, the feature of each face node is initialized based on the number of edges it contains, while the weight of each edge between two connected face nodes is determined by the number of edges they share ($\mathbf{FeF}_i / M_e$). These initial node features and edge weights are input into a two-layer GCN, which aggregates and transforms information from neighboring nodes. This process yields a final embedding matrix with shape $N_f^i \times d_{ev}$, where $d_{ev}$ is the dimensionality of the aggregated features for each face. The resulting face node embeddings are then combined with the face ID embeddings to form the final aggregated face features. It is worth noting that this face feature extraction step is optional, as it may slightly increase training time while offering marginal performance gains.

\begin{figure}
	\centering
	\includegraphics[width=0.6\linewidth]{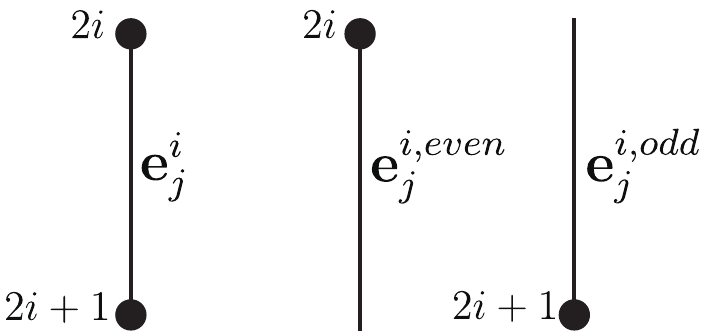}
	\caption{Each edge in the $\mathbf{E}_i$ is duplicated to form $\Tilde{\mathbf{E}}_i$, with even- and odd-indexed edges assigned to represent its two endpoints.} 
	\label{fig: even-odd} 
\end{figure}

\textbf{Endpoint embedding.} To generate the endpoint embeddings, each edge in the $\mathbf{E}_i$ is duplicated, producing two copies corresponding to its two endpoints, as illustrated in \cref{fig: even-odd}. The resulting duplicated edges is denoted as
\begin{equation} 
	\Tilde{\mathbf{E}}_i = \left\{ \mathbf{e}_0^{i,even}, \mathbf{e}_0^{i,odd}, \mathbf{e}_1^{i,even}, \cdots, \mathbf{e}_{N_e^i-1}^{i,odd} \right\},
\end{equation}
To encode the endpoint-specific information, we introduce two learnable vectors, $emb_{even} \in \mathbb{R}^{d_{ev}}$ and $emb_{odd} \in \mathbb{R}^{d_{ev}}$, representing the embeddings for ``even-indexed edges'' and ``odd-indexed edges'', respectively. The final endpoint embedding for the edge set $\Tilde{\mathbf{E}}i$ is constructed as a sequence of these embeddings, defined as
\begin{equation} 
	EDP\left(\Tilde{\mathbf{E}}_i\right) = \left\{ emb_{even}, emb_{odd}, \cdots, emb_{even}, emb_{odd} \right\},
\end{equation}
where $\left|EDP\left(\Tilde{\mathbf{E}}_i\right)\right|= 2N_e^i$. This structured embedding provides a consistent and learnable representation for edge endpoints, enabling efficient feature encoding for embedding tasks.

\textbf{Training workflow.} During the training phase, the edges $\mathbf{E}_i$ together with two special tokens $\mathbf{e}_{loop}$ and $\mathbf{e}_{face}$ are embedded into $2N_e^i+2$ vectors, each with dimension $d_{ev}$. These embeddings are processed by the Transformer encoder to obtain contextual embeddings $\mathcal{E}_\theta^{EV}(\mathbf{E}_i) \in \mathbb{R}^{(2N_e^i+2)\times d_{ev}}$. For the decoding process, each integer in the ground-truth edge-vertex sequence $\mathbf{EV}_i^{seq}$ is mapped to its corresponding contextual embedding, maintaining the edge connectivity order. These aligned embeddings, combined with the positional embedding $POS_\theta$, are then processed by a masked Transformer decoder followed by a pointer network to produce decoder embeddings $\mathcal{D}_\theta^{EV}(\mathcal{E}_\theta^{EV}(\mathbf{E}_i),\mathbf{EV}_i^{seq}) \in \mathbb{R}^{|\mathbf{EV}_i^{seq}| \times d_{ev}}$. The pointer attention scores, computed as:
\begin{equation}
	\text{Pointer}(\mathbf{E}_i, \mathbf{EV}_i^{seq}) = \mathcal{D}_\theta^{EV}(\mathcal{E}_\theta^{EV}(\mathbf{E}_i),\mathbf{EV}_i^{seq}) \cdot \mathcal{E}_\theta^{EV}(\mathbf{E}_i)^T ,
\end{equation}
represent the probability distribution over all possible edges at each position after softmax normalization. \cref{tab: ev-shape} summarizes the output dimensions of each module. 

\begin{table*}
	\centering
	\begin{tabular}{ccccc}
		\toprule
		& Input & Embedding & (Masked) Transformer blocks & Output layer \\ 
		\midrule
		Encoder & $N_e^i$  & $(2N_e^i+2) \times d_{ev}$ & $(2N_e^i+2) \times d_{ev}$   &  -  \\ 
		Decoder & $(2N_e^i+2) \times d_{ev}, |\mathbf{EV}_i^{seq}|$ & $|\mathbf{EV}_i^{seq}| \times d_{ev}$  & $|\mathbf{EV}_i^{seq}| \times d_{ev}$  & $|\mathbf{EV}_i^{seq}| \times (2N_e^i+2)$  \\ 
		\bottomrule
	\end{tabular}
	\caption{Summary of output shapes for each module of our edge-vertex model during the training phase.}
	\label{tab: ev-shape}
\end{table*}

\subsection{Details of geometry generation network}
\textbf{Topology-aware geometric generation.} For the face bounding box generation, we introduce a learnable matrix in $\mathbb{R}^{(M_e+1)\times d_{geom}}$ to encode shared-edge relationships between faces, where $d_{geom}$ denotes the embedding dimension. The attention weights between faces are adjusted by adding the corresponding edge embeddings based on the number of shared edges. In vertex coordinate generation, we incorporate edge connectivity information through two learnable embeddings that indicate whether vertices are connected. These embeddings modulate the attention weights in the Transformer. Additionally, we augment each vertex's input representation by incorporating the average embedding of its adjacent faces' bounding boxes. For edge and face geometry generation, unlike the previous stages, we maintain the original attention mechanism without additional adjustment. Instead, we enhance the input token embeddings with topological information. Specifically, for edge geometry generation, we enrich the input token embeddings with: (1) Bounding box embeddings of faces containing the edge. (2) Coordinate embeddings of the edge's two endpoint vertices. Similarly, for face geometry generation, the input token embeddings are enhanced with: (1) The corresponding face's bounding box embedding. (2) Coordinate embeddings of the face's vertices. (3) Geometric embeddings of the face's boundary edges. This hierarchical approach ensures that each generation stage benefits from both the topological structure and previously generated geometric information, leading to more coherent and physically valid results. 

\textbf{Learning of B-spline representation.} We adopt B-spline representations for both edge and face geometries. For edges, we employ cubic B-splines with knot vector $\{0,0,0,0,1,1,1,1\}$, while faces are represented using bi-cubic B-splines with knot vectors $\{0,0,0,0,1,1,1,1\}$ in both $u$ and $v$ directions. This B-spline representation exhibits inherent symmetry, meaning different orderings of the same control points can produce identical geometric forms. To ensure consistency in our training data, we utilize OpenCascade's curve and surface conversion functions \cite{pyOCCT} to standardize the B-spline representations. Specifically, all edges and faces in the dataset are converted to B-spline form, establishing a consistent ordering of control points that is maintained throughout the training process. During inference, we generate the control points sequentially using our trained model and construct the B-spline geometries through OpenCascade's built-in functions, following the same control point ordering convention established during training. \cref{fig: ctrs} illustrates examples of edge and face geometries in B-spline representation along with their corresponding control points.

\begin{figure}
	\centering
	\begin{subfigure}[b]{0.45\linewidth}
		\centering
		\includegraphics[width=0.7\linewidth]{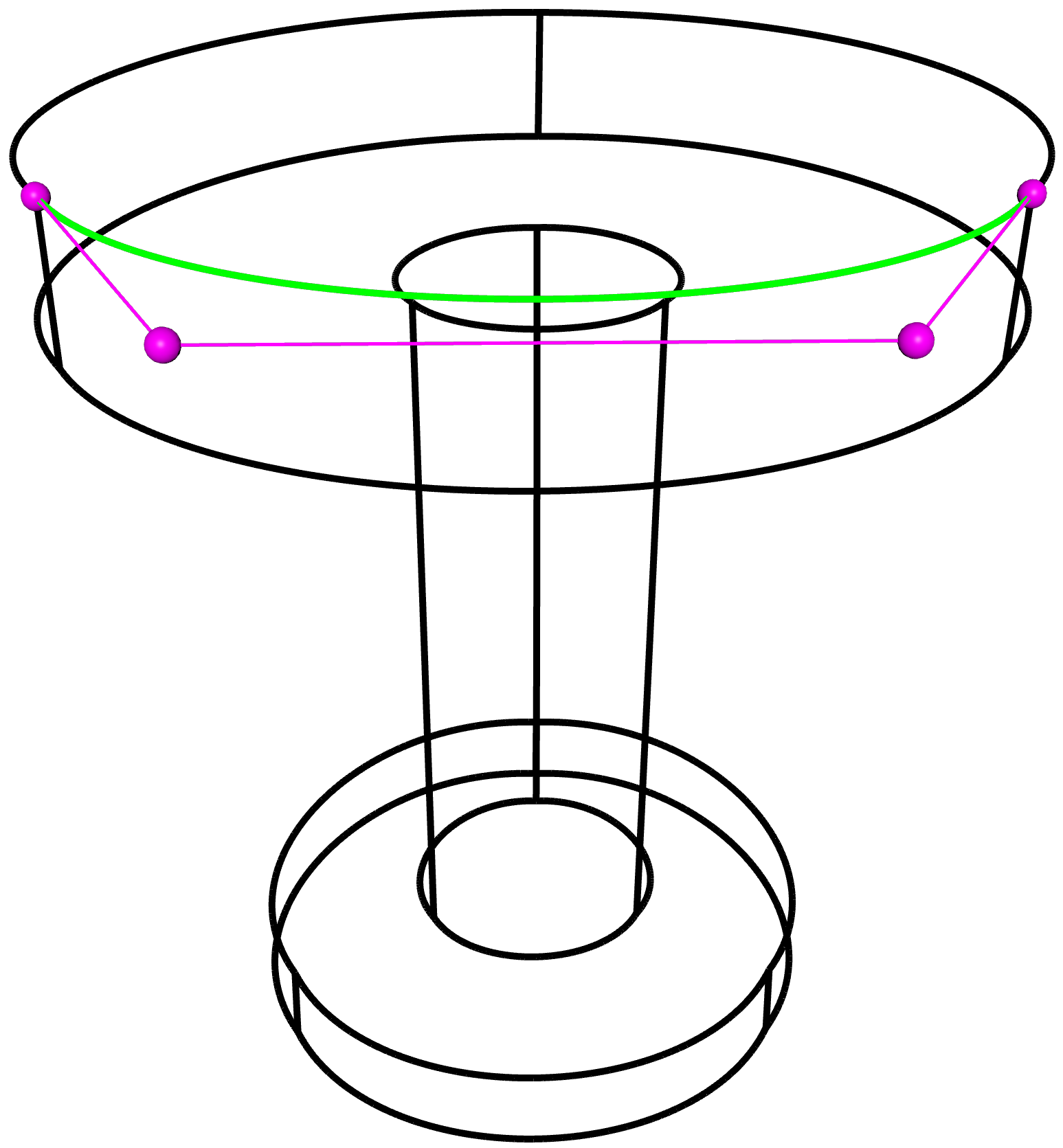} 
	\end{subfigure}
	\hfill 
	\begin{subfigure}[b]{0.45\linewidth}
		\centering
		\includegraphics[width=0.7\linewidth]{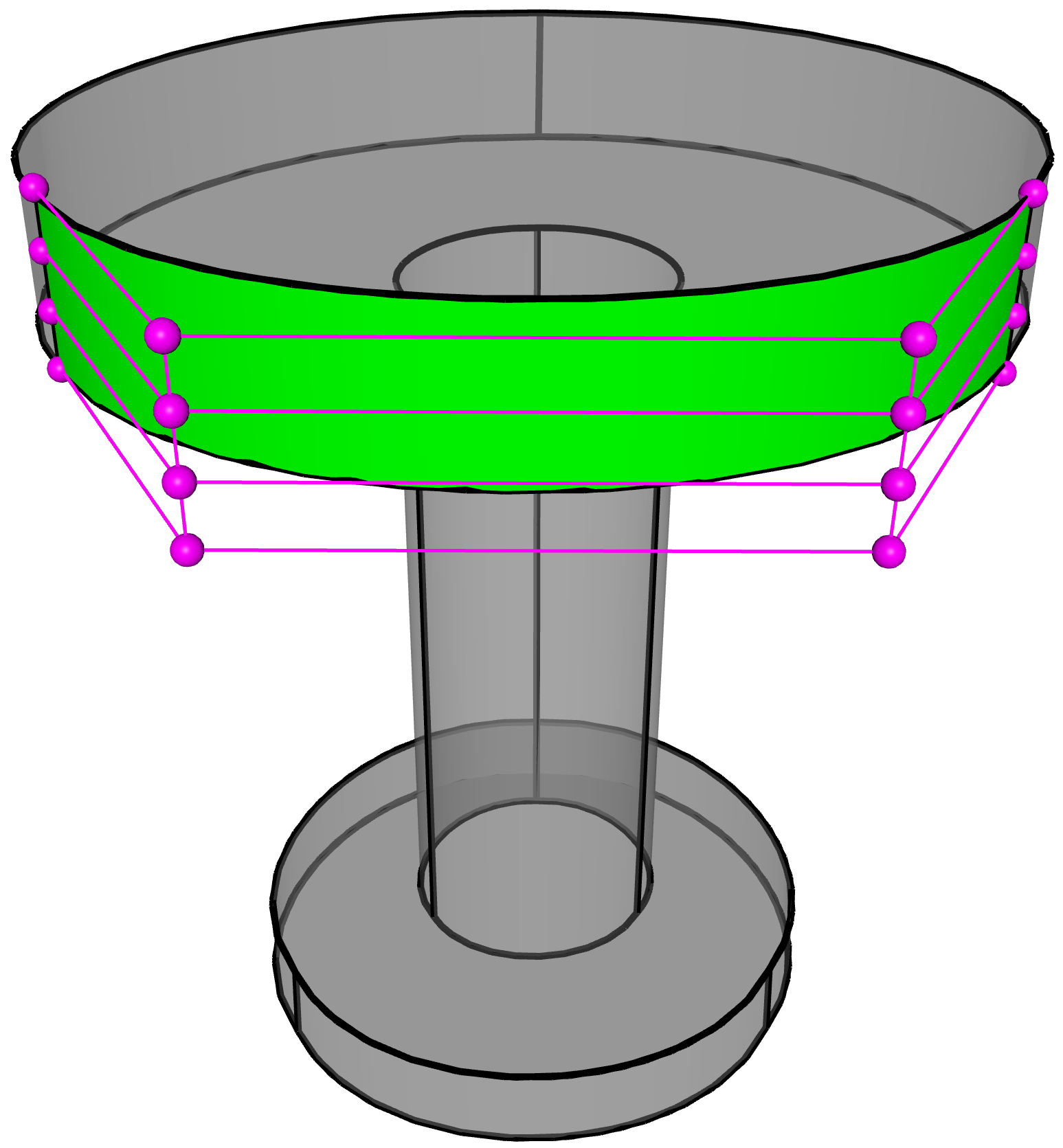} 
	\end{subfigure}
	\caption{Visualization of B-spline representations in our framework. Each edge is represented by a cubic B-spline with $4$ control points, while each face is parameterized by a bi-cubic B-spline surface with a $4\times 4$ control point grid.}
	\label{fig: ctrs}
\end{figure}

\section{Experiments}
\begin{figure*}
	\centering
	\includegraphics[width=0.95\linewidth]{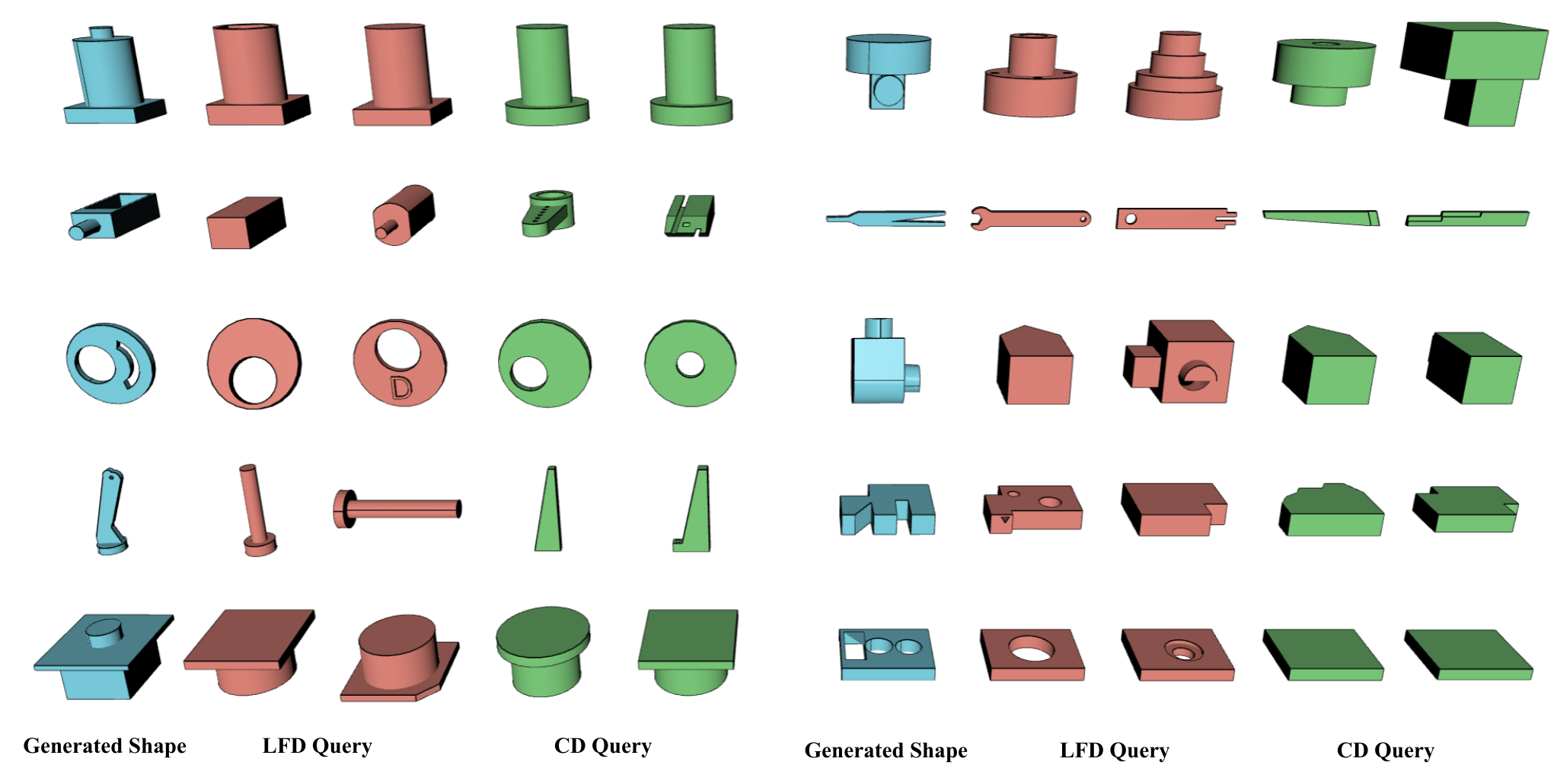}
	\caption{Representative examples of generated shapes alongside their two nearest neighbors from the training set, retrieved using Chamfer Distance and Light Field Distance metrics. The distinct geometric differences highlight the novel nature of the generated shapes.} 
	\label{fig: novelty} 
\end{figure*}

\subsection{Implementation details}
We implement our framework in PyTorch and conduct all experiments on 4 NVIDIA A800 GPUs. The B-rep dataset is randomly split into training ($90\%$), validation ($5\%$), and test ($5\%$) sets. The transformer embedding dimensions are set to $d_{ef}=128$ for the edge-face model, $d_{ev}=256$ for the edge-vertex model, and $d_{geom}=512$ for the geometry generation models. To accommodate the topology structure, we set the maximum number of shared edges between faces to $M_e=5$ and the maximum number of faces to $M_f=30$. For topology learning, we train the edge-face model for $2,000$ epochs and the edge-vertex model for $1,000$ epochs. The four geometry generation models are each trained for $3,000$ epochs. All models are optimized using Adam with an initial learning rate of $5\times10^{-4}$ ($1\times 10^{-4}$ for the ABC dataset) and weight decay of $1\times10^{-6}$, with a batch size of 512. For the geometry generation models, we employ a linear-schedule DDPM \cite{ho2020denoising} with $1,000$ diffusion steps during training. The noise schedule follows a linear beta schedule from $1\times10^{-4}$ to $2\times10^{-2}$. 

\subsection{Shape retrieval}
To further evaluate DTGBrepGen's capability in generating novel shapes rather than merely memorizing training samples, we conduct a shape retrieval experiment \cite{hui2022wavelet} between the generated samples and the training dataset.. Following established protocols, we compute both Chamfer Distance (CD) and Light Field Distance (LFD) \cite{chen2003visual} between 500 randomly generated shapes and the entire training set. As shown in \cref{fig: novelty}, we visualize representative examples of our generated shapes alongside their two most similar counterparts from the training set retrieved using both metrics. The distinct geometric variations between generated samples and their nearest neighbors in the training set demonstrate that DTGBrepGen is not simply reproducing training examples. These results collectively validate that our topology-geometry decoupled approach enables the creation of novel, yet realistic B-rep models that extend beyond the training distribution while maintaining high geometric quality.

\subsection{Qualitative results of the ablation study}
\begin{figure*}
	\centering
	\begin{subfigure}[b]{0.49\linewidth}
		\centering
		\includegraphics[width=\linewidth]{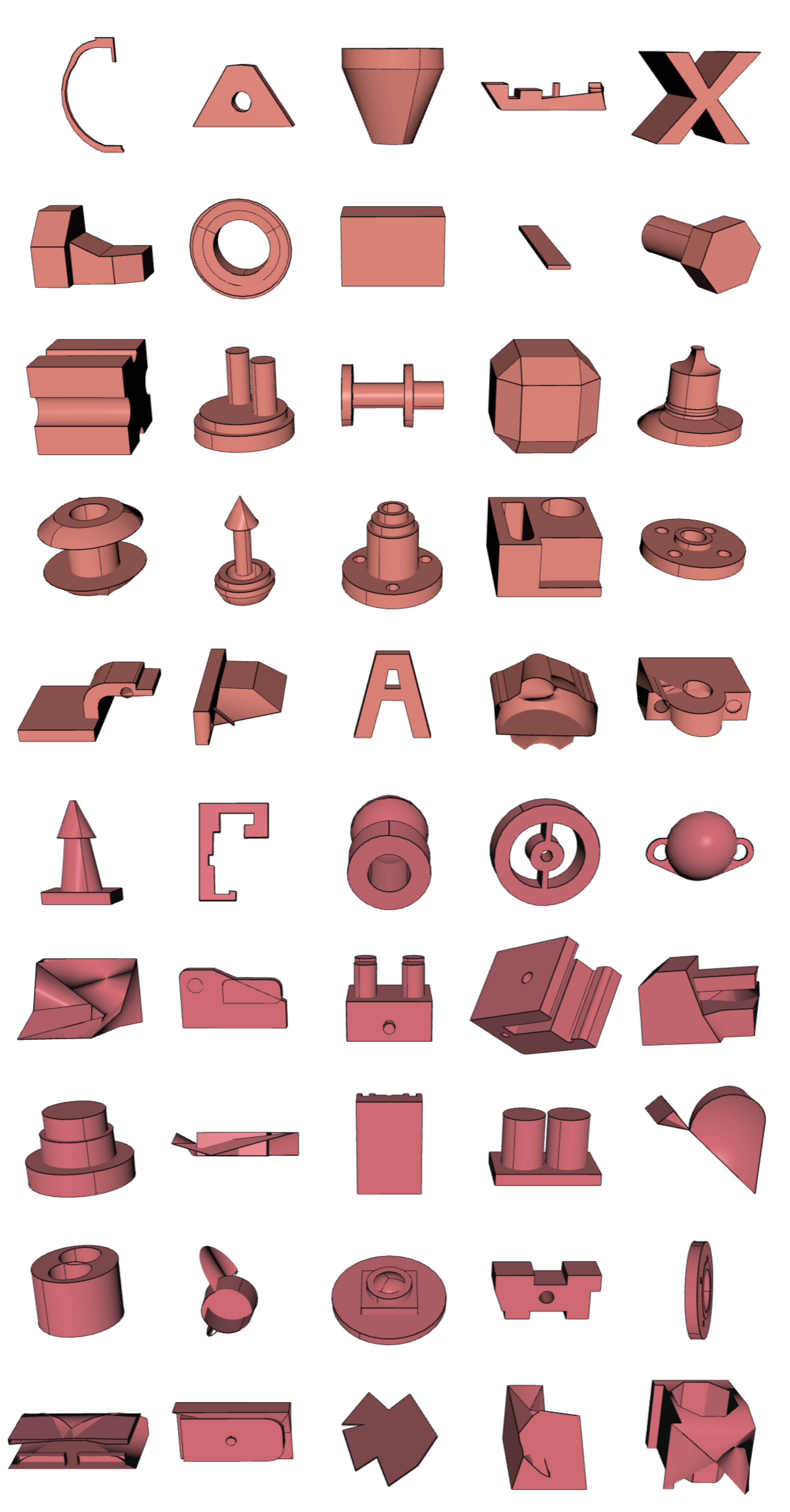} 
		\caption{Point-based approach}
	\end{subfigure}
	\hfill 
	\begin{subfigure}[b]{0.49\linewidth}
		\centering
		\includegraphics[width=\linewidth]{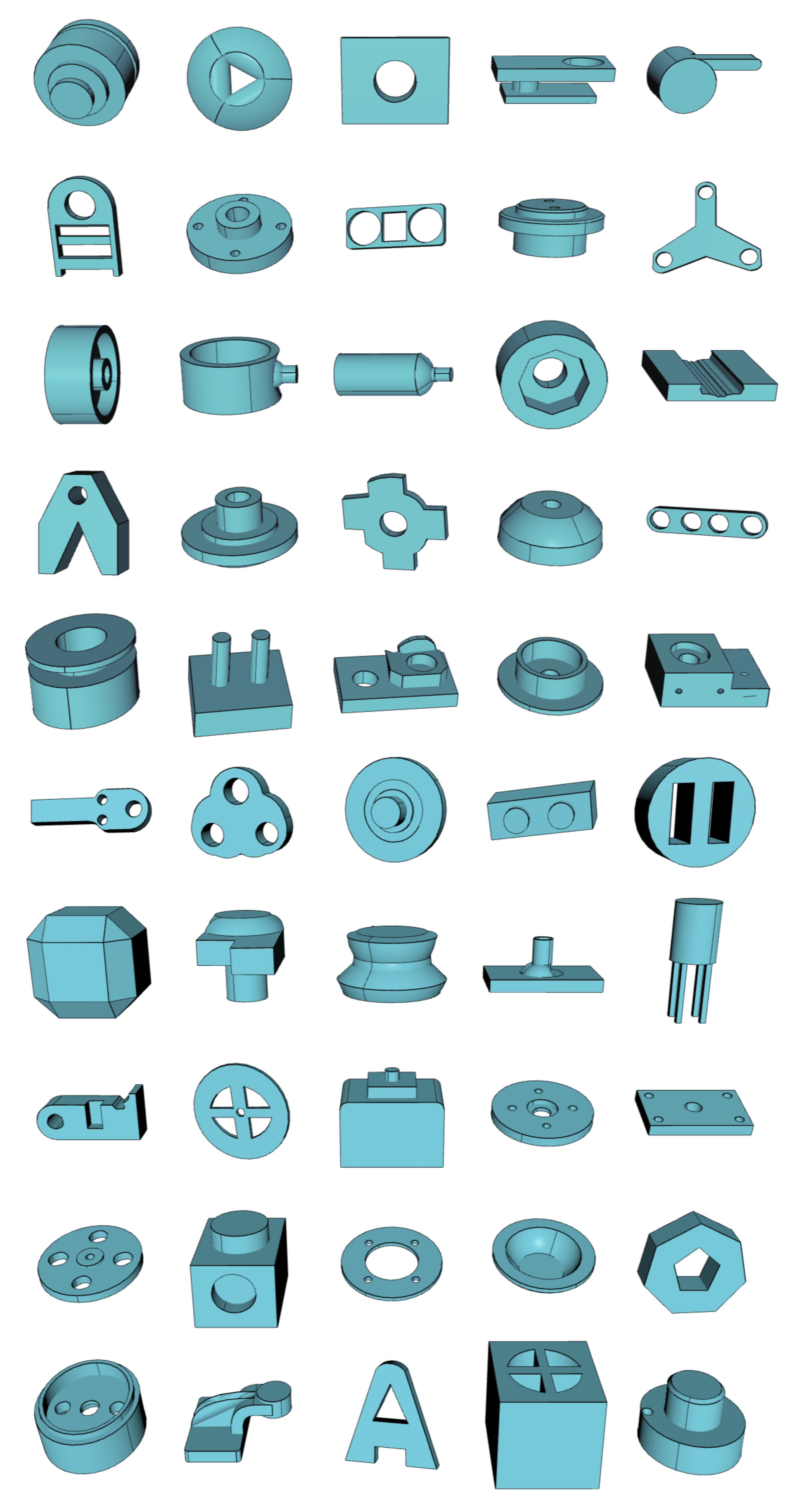} 
		\caption{B-spline-based approach (ours)}
	\end{subfigure}
	\caption{Qualitative comparison between B-spline-based and point-based geometric representations on the ABC dataset.}
	\label{fig: abc}
\end{figure*}

To demonstrate the advantages of B-spline representation over discrete point-based approaches, we conduct extensive qualitative comparisons on the ABC dataset, as illustrated in \cref{fig: abc}. Our B-spline-based method exhibits consistently superior performance, producing smoother and more geometrically accurate details compared to the point-based variant. These qualitative results, aligned with our quantitative findings in the main paper, further validate the effectiveness of directly learning B-spline control point distributions over discrete point-based representations. The superior performance can be attributed to the B-spline representation's inherent ability to capture continuous geometric features with fewer parameters, leading to more robust and accurate shape generation.

\subsection{More examples generated by DTGBrepGen}
\begin{figure*}
	\centering
	\includegraphics[width=0.9\linewidth]{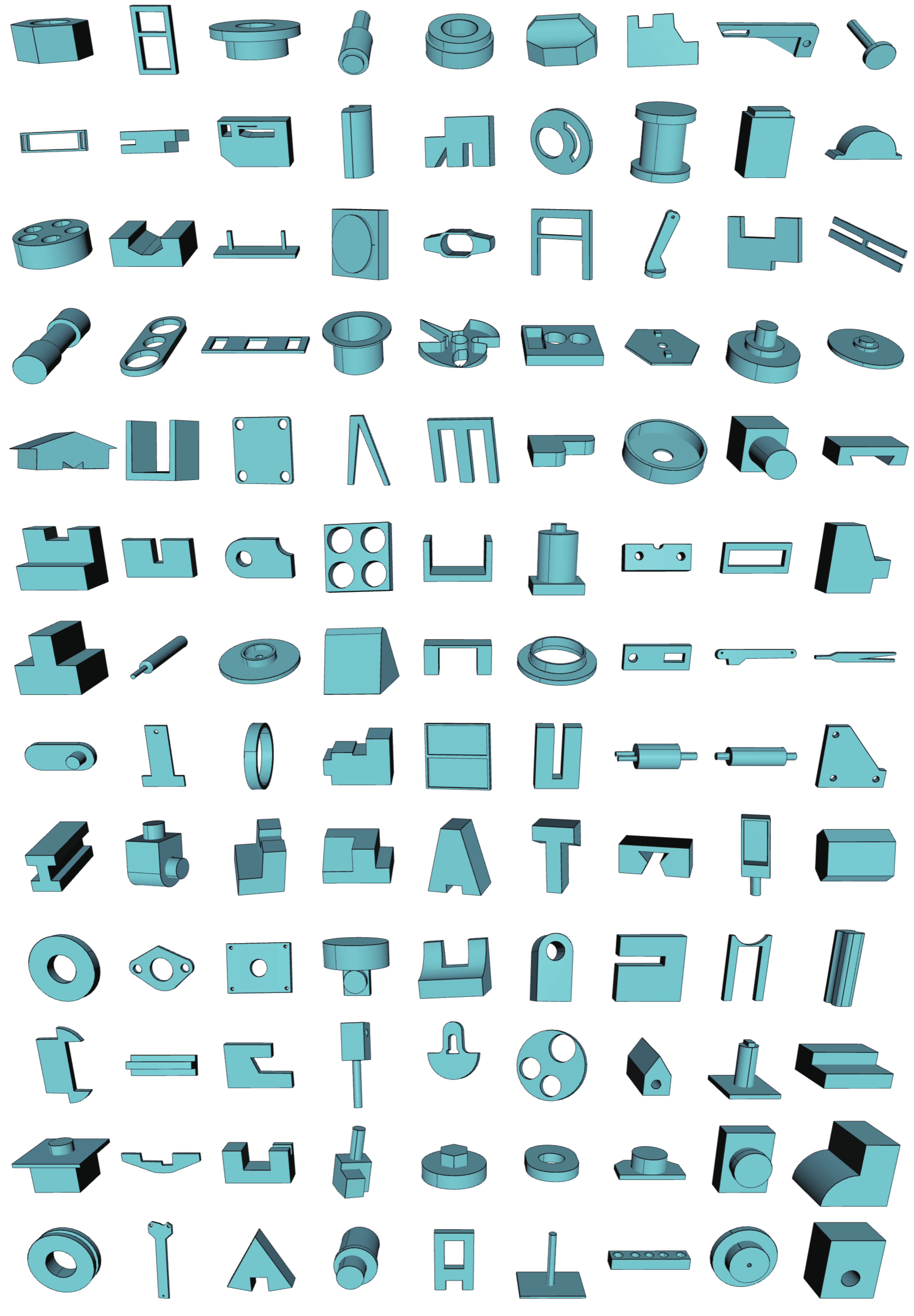}
	\caption{Examples of B-rep models generated by DTGBrepGen on the DeepCAD dataset.} 
	\label{fig: deepcad} 
\end{figure*}

\begin{figure*}
	\centering
	\includegraphics[width=0.87\linewidth]{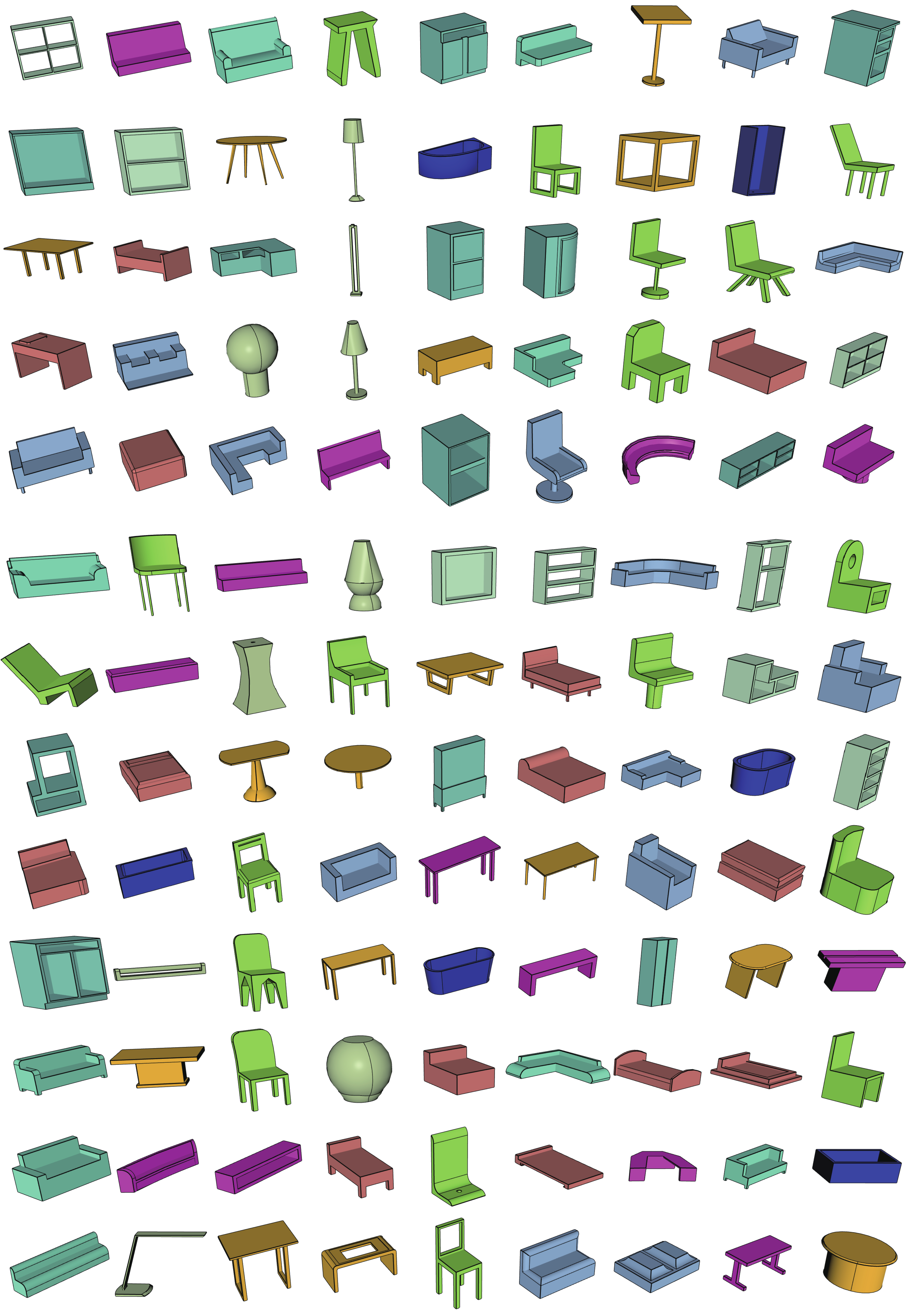}
	\caption{Examples of class-conditioned generation on the Furniture dataset, with distinct colors representing different categories.} 
	\label{fig: furniture} 
\end{figure*}

We present additional generation results to demonstrate DTGBrepGen's versatility. \cref{fig: deepcad} shows diverse unconditional generation examples on the DeepCAD dataset. \cref{fig: furniture} presents class-conditioned generation results across different furniture categories. \cref{fig: point2brep} illustrates our model's capability in translating point clouds to B-rep models.

\onecolumn
\begin{figure}
	\centering
	\includegraphics[width=0.9\linewidth]{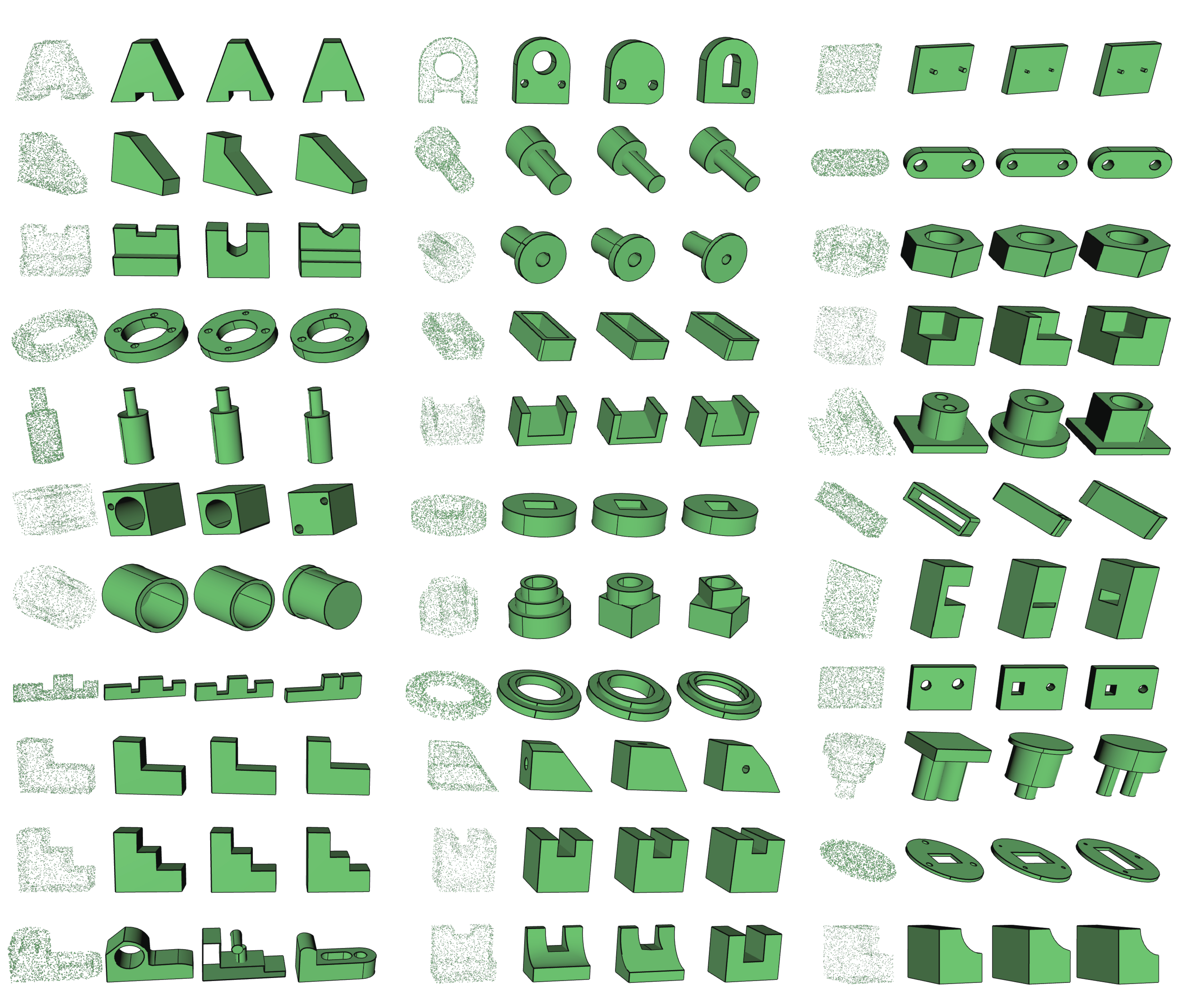}
	\caption{Examples of point cloud-conditioned B-rep generation on the DeepCAD dataset.} 
	\label{fig: point2brep} 
\end{figure}
\twocolumn

\end{document}


%% file: main.bbl
\begin{thebibliography}{38}
\providecommand{\natexlab}[1]{#1}
\providecommand{\url}[1]{\texttt{#1}}
\expandafter\ifx\csname urlstyle\endcsname\relax
  \providecommand{\doi}[1]{doi: #1}\else
  \providecommand{\doi}{doi: \begingroup \urlstyle{rm}\Url}\fi

\bibitem[Benk{\H{o}} and V{\'a}rady(2004)]{benkHo2004segmentation}
P{\'a}l Benk{\H{o}} and Tam{\'a}s V{\'a}rady.
\newblock Segmentation methods for smooth point regions of conventional
  engineering objects.
\newblock \emph{Computer-Aided Design}, 36\penalty0 (6):\penalty0 511--523,
  2004.

\bibitem[Cao et~al.(2020)Cao, Robinson, Hua, Boussuge, Colligan, and
  Pan]{cao2020graph}
Weijuan Cao, Trevor Robinson, Yang Hua, Flavien Boussuge, Andrew~R Colligan,
  and Wanbin Pan.
\newblock Graph representation of 3d cad models for machining feature
  recognition with deep learning.
\newblock In \emph{International Design Engineering Technical Conferences and
  Computers and Information in Engineering Conference}, page V11AT11A003.
  American Society of Mechanical Engineers, 2020.

\bibitem[Chen et~al.(2003)Chen, Tian, Shen, and Ouhyoung]{chen2003visual}
Ding-Yun Chen, Xiao-Pei Tian, Yu-Te Shen, and Ming Ouhyoung.
\newblock On visual similarity based 3d model retrieval.
\newblock In \emph{Computer graphics forum}, pages 223--232. Wiley Online
  Library, 2003.

\bibitem[Dhariwal and Nichol(2021)]{dhariwal2021diffusion}
Prafulla Dhariwal and Alexander Nichol.
\newblock Diffusion models beat gans on image synthesis.
\newblock \emph{Advances in Neural Information Processing Systems},
  34:\penalty0 8780--8794, 2021.

\bibitem[Ellis et~al.(2019)Ellis, Nye, Pu, Sosa, Tenenbaum, and
  Solar-Lezama]{ellis2019write}
Kevin Ellis, Maxwell Nye, Yewen Pu, Felix Sosa, Josh Tenenbaum, and Armando
  Solar-Lezama.
\newblock Write, execute, assess: Program synthesis with a repl.
\newblock \emph{Advances in Neural Information Processing Systems}, 32, 2019.

\bibitem[Goodfellow et~al.(2014)Goodfellow, Pouget-Abadie, Mirza, Xu,
  Warde-Farley, Ozair, Courville, and Bengio]{goodfellow2014generative}
Ian Goodfellow, Jean Pouget-Abadie, Mehdi Mirza, Bing Xu, David Warde-Farley,
  Sherjil Ozair, Aaron Courville, and Yoshua Bengio.
\newblock Generative adversarial nets.
\newblock \emph{Advances in Neural Information Processing Systems}, 27, 2014.

\bibitem[Guo et~al.(2022)Guo, Liu, Pan, Liu, Tong, and Guo]{guo2022complexgen}
Haoxiang Guo, Shilin Liu, Hao Pan, Yang Liu, Xin Tong, and Baining Guo.
\newblock Complexgen: Cad reconstruction by b-rep chain complex generation.
\newblock \emph{ACM Transactions on Graphics (TOG)}, 41\penalty0 (4):\penalty0
  1--18, 2022.

\bibitem[Ho and Salimans(2022)]{ho2022classifier}
Jonathan Ho and Tim Salimans.
\newblock Classifier-free diffusion guidance.
\newblock \emph{arXiv preprint arXiv:2207.12598}, 2022.

\bibitem[Ho et~al.(2020)Ho, Jain, and Abbeel]{ho2020denoising}
Jonathan Ho, Ajay Jain, and Pieter Abbeel.
\newblock Denoising diffusion probabilistic models.
\newblock \emph{Advances in Neural Information Processing Systems},
  33:\penalty0 6840--6851, 2020.

\bibitem[Hui et~al.(2022)Hui, Li, Hu, and Fu]{hui2022wavelet}
Ka-Hei Hui, Ruihui Li, Jingyu Hu, and Chi-Wing Fu.
\newblock Neural wavelet-domain diffusion for 3d shape generation.
\newblock 2022.

\bibitem[Jayaraman et~al.(2021)Jayaraman, Sanghi, Lambourne, Willis, Davies,
  Shayani, and Morris]{jayaraman2021uv}
Pradeep~Kumar Jayaraman, Aditya Sanghi, Joseph~G Lambourne, Karl~DD Willis,
  Thomas Davies, Hooman Shayani, and Nigel Morris.
\newblock Uv-net: Learning from boundary representations.
\newblock In \emph{Proceedings of the IEEE/CVF Conference on Computer Vision
  and Pattern Recognition}, pages 11703--11712, 2021.

\bibitem[Jayaraman et~al.(2023)Jayaraman, Lambourne, Desai, Willis, Sanghi, and
  Morris]{jayaraman2023solidgen}
Pradeep~Kumar Jayaraman, Joseph~George Lambourne, Nishkrit Desai, Karl Willis,
  Aditya Sanghi, and Nigel J.~W. Morris.
\newblock Solidgen: An autoregressive model for direct b-rep synthesis.
\newblock \emph{Transactions on Machine Learning Research}, 2023.

\bibitem[Kingma and Welling(2014)]{DBLP:journals/corr/KingmaW13}
Diederik~P. Kingma and Max Welling.
\newblock Auto-encoding variational bayes.
\newblock In \emph{2nd International Conference on Learning Representations,
  {ICLR} 2014, Banff, AB, Canada, April 14-16, 2014, Conference Track
  Proceedings}, 2014.

\bibitem[Kipf and Welling(2017)]{DBLP:conf/iclr/KipfW17}
Thomas~N. Kipf and Max Welling.
\newblock Semi-supervised classification with graph convolutional networks.
\newblock In \emph{5th International Conference on Learning Representations,
  {ICLR} 2017, Toulon, France, April 24-26, 2017, Conference Track
  Proceedings}, 2017.

\bibitem[Koch et~al.(2019)Koch, Matveev, Jiang, Williams, Artemov, Burnaev,
  Alexa, Zorin, and Panozzo]{koch2019abc}
Sebastian Koch, Albert Matveev, Zhongshi Jiang, Francis Williams, Alexey
  Artemov, Evgeny Burnaev, Marc Alexa, Denis Zorin, and Daniele Panozzo.
\newblock Abc: A big cad model dataset for geometric deep learning.
\newblock In \emph{Proceedings of the IEEE/CVF Conference on Computer Vision
  and Pattern Recognition}, pages 9601--9611, 2019.

\bibitem[Lambourne et~al.(2022)Lambourne, Willis, Jayaraman, Zhang, Sanghi, and
  Malekshan]{lambourne2022reconstructing}
Joseph~George Lambourne, Karl Willis, Pradeep~Kumar Jayaraman, Longfei Zhang,
  Aditya Sanghi, and Kamal~Rahimi Malekshan.
\newblock Reconstructing editable prismatic cad from rounded voxel models.
\newblock In \emph{SIGGRAPH Asia 2022 Conference Papers}, pages 1--9, 2022.

\bibitem[Laughlin(2020)]{pyOCCT}
Trevor Laughlin.
\newblock pyocct -- python bindings for opencascade via pybind11, 2020.
\newblock https://github.com/trelau/pyOCCT.

\bibitem[Li et~al.(2023)Li, Guo, Zhang, and Yan]{li2023secad}
Pu Li, Jianwei Guo, Xiaopeng Zhang, and Dong-Ming Yan.
\newblock Secad-net: Self-supervised cad reconstruction by learning
  sketch-extrude operations.
\newblock In \emph{Proceedings of the IEEE/CVF Conference on Computer Vision
  and Pattern Recognition}, pages 16816--16826, 2023.

\bibitem[Liu et~al.(2024)Liu, Obukhov, Wegner, and Schindler]{liu2024point2cad}
Yujia Liu, Anton Obukhov, Jan~Dirk Wegner, and Konrad Schindler.
\newblock Point2cad: Reverse engineering cad models from 3d point clouds.
\newblock In \emph{Proceedings of the IEEE/CVF Conference on Computer Vision
  and Pattern Recognition}, pages 3763--3772, 2024.

\bibitem[Nash et~al.(2020)Nash, Ganin, Eslami, and Battaglia]{nash2020polygen}
Charlie Nash, Yaroslav Ganin, SM~Ali Eslami, and Peter Battaglia.
\newblock Polygen: An autoregressive generative model of 3d meshes.
\newblock In \emph{International Conference on Machine Learning}, pages
  7220--7229. PMLR, 2020.

\bibitem[Qi et~al.(2017)Qi, Yi, Su, and Guibas]{qi2017pointnet++}
Charles~Ruizhongtai Qi, Li Yi, Hao Su, and Leonidas~J Guibas.
\newblock Pointnet++: Deep hierarchical feature learning on point sets in a
  metric space.
\newblock \emph{Advances in Neural Information Processing Systems}, 30, 2017.

\bibitem[Ritchie et~al.(2023)Ritchie, Guerrero, Jones, Mitra, Schulz, Willis,
  and Wu]{ritchie2023neurosymbolic}
Daniel Ritchie, Paul Guerrero, R~Kenny Jones, Niloy~J Mitra, Adriana Schulz,
  Karl~DD Willis, and Jiajun Wu.
\newblock Neurosymbolic models for computer graphics.
\newblock In \emph{Computer Graphics Forum}, pages 545--568. Wiley Online
  Library, 2023.

\bibitem[Smirnov et~al.(2021)Smirnov, Bessmeltsev, and
  Solomon]{smirnov2021learning}
Dmitriy Smirnov, Mikhail Bessmeltsev, and Justin Solomon.
\newblock Learning manifold patch-based representations of man-made shapes.
\newblock In \emph{International Conference on Learning Representations}, 2021.

\bibitem[Tian et~al.(2019)Tian, Luo, Sun, Ellis, Freeman, Tenenbaum, and
  Wu]{tian2018learning}
Yonglong Tian, Andrew Luo, Xingyuan Sun, Kevin Ellis, William~T. Freeman,
  Joshua~B. Tenenbaum, and Jiajun Wu.
\newblock Learning to infer and execute 3d shape programs.
\newblock In \emph{International Conference on Learning Representations}, 2019.

\bibitem[Uy et~al.(2022)Uy, Chang, Sung, Goel, Lambourne, Birdal, and
  Guibas]{uy2022point2cyl}
Mikaela~Angelina Uy, Yen-Yu Chang, Minhyuk Sung, Purvi Goel, Joseph~G
  Lambourne, Tolga Birdal, and Leonidas~J Guibas.
\newblock Point2cyl: Reverse engineering 3d objects from point clouds to
  extrusion cylinders.
\newblock In \emph{Proceedings of the IEEE/CVF Conference on Computer Vision
  and Pattern Recognition}, pages 11850--11860, 2022.

\bibitem[Vaswani(2017)]{vaswani2017attention}
A Vaswani.
\newblock Attention is all you need.
\newblock \emph{Advances in Neural Information Processing Systems}, 2017.

\bibitem[Vignac et~al.(2023)Vignac, Krawczuk, Siraudin, Wang, Cevher, and
  Frossard]{vignac2023digress}
Clement Vignac, Igor Krawczuk, Antoine Siraudin, Bohan Wang, Volkan Cevher, and
  Pascal Frossard.
\newblock Digress: Discrete denoising diffusion for graph generation.
\newblock In \emph{The Eleventh International Conference on Learning
  Representations}, 2023.

\bibitem[Vinyals et~al.(2015)Vinyals, Fortunato, and
  Jaitly]{vinyals2015pointer}
Oriol Vinyals, Meire Fortunato, and Navdeep Jaitly.
\newblock Pointer networks.
\newblock \emph{Advances in Neural Information Processing Systems}, 28, 2015.

\bibitem[Wang et~al.(2020)Wang, Xu, Xu, Tagliasacchi, Zhou, Mahdavi-Amiri, and
  Zhang]{wang2020pie}
Xiaogang Wang, Yuelang Xu, Kai Xu, Andrea Tagliasacchi, Bin Zhou, Ali
  Mahdavi-Amiri, and Hao Zhang.
\newblock Pie-net: Parametric inference of point cloud edges.
\newblock \emph{Advances in Neural Information Processing Systems},
  33:\penalty0 20167--20178, 2020.

\bibitem[Weiler(1986)]{weiler1986topological}
Kevin~J Weiler.
\newblock \emph{Topological structures for geometric modeling (Boundary
  representation, manifold, radial edge structure)}.
\newblock Rensselaer Polytechnic Institute, 1986.

\bibitem[Williams and Zipser(1989)]{williams1989learning}
Ronald~J Williams and David Zipser.
\newblock A learning algorithm for continually running fully recurrent neural
  networks.
\newblock \emph{Neural Computation}, 1\penalty0 (2):\penalty0 270--280, 1989.

\bibitem[Willis et~al.(2021)Willis, Pu, Luo, Chu, Du, Lambourne, Solar-Lezama,
  and Matusik]{willis2021fusion}
Karl~DD Willis, Yewen Pu, Jieliang Luo, Hang Chu, Tao Du, Joseph~G Lambourne,
  Armando Solar-Lezama, and Wojciech Matusik.
\newblock Fusion 360 gallery: A dataset and environment for programmatic cad
  construction from human design sequences.
\newblock \emph{ACM Transactions on Graphics (TOG)}, 40\penalty0 (4):\penalty0
  1--24, 2021.

\bibitem[Willis et~al.(2022)Willis, Jayaraman, Chu, Tian, Li, Grandi, Sanghi,
  Tran, Lambourne, Solar-Lezama, et~al.]{willis2022joinable}
Karl~DD Willis, Pradeep~Kumar Jayaraman, Hang Chu, Yunsheng Tian, Yifei Li,
  Daniele Grandi, Aditya Sanghi, Linh Tran, Joseph~G Lambourne, Armando
  Solar-Lezama, et~al.
\newblock Joinable: Learning bottom-up assembly of parametric cad joints.
\newblock In \emph{Proceedings of the IEEE/CVF Conference on Computer Vision
  and Pattern Recognition}, pages 15849--15860, 2022.

\bibitem[Wu et~al.(2021)Wu, Xiao, and Zheng]{wu2021deepcad}
Rundi Wu, Chang Xiao, and Changxi Zheng.
\newblock Deepcad: A deep generative network for computer-aided design models.
\newblock In \emph{Proceedings of the IEEE/CVF International Conference on
  Computer Vision}, pages 6772--6782, 2021.

\bibitem[Xu et~al.(2022)Xu, Willis, Lambourne, Cheng, Jayaraman, and
  Furukawa]{xu2022skexgen}
Xiang Xu, Karl~DD Willis, Joseph~G Lambourne, Chin-Yi Cheng, Pradeep~Kumar
  Jayaraman, and Yasutaka Furukawa.
\newblock Skexgen: Autoregressive generation of cad construction sequences with
  disentangled codebooks.
\newblock In \emph{International Conference on Machine Learning}, pages
  24698--24724. PMLR, 2022.

\bibitem[Xu et~al.(2023)Xu, Jayaraman, Lambourne, Willis, and
  Furukawa]{xu2023hierarchical}
Xiang Xu, Pradeep~Kumar Jayaraman, Joseph~G. Lambourne, Karl~D.D. Willis, and
  Yasutaka Furukawa.
\newblock Hierarchical neural coding for controllable cad model generation.
\newblock In \emph{Proceedings of the 40th International Conference on Machine
  Learning}, 2023.

\bibitem[Xu et~al.(2024)Xu, Lambourne, Jayaraman, Wang, Willis, and
  Furukawa]{xu2024brepgen}
Xiang Xu, Joseph Lambourne, Pradeep Jayaraman, Zhengqing Wang, Karl Willis, and
  Yasutaka Furukawa.
\newblock Brepgen: A b-rep generative diffusion model with structured latent
  geometry.
\newblock \emph{ACM Transactions on Graphics (TOG)}, 43\penalty0 (4):\penalty0
  1--14, 2024.

\bibitem[Ying et~al.(2021)Ying, Cai, Luo, Zheng, Ke, He, Shen, and
  Liu]{ying2021transformers}
Chengxuan Ying, Tianle Cai, Shengjie Luo, Shuxin Zheng, Guolin Ke, Di He,
  Yanming Shen, and Tie-Yan Liu.
\newblock Do transformers really perform badly for graph representation?
\newblock \emph{Advances in Neural Information Processing Systems},
  34:\penalty0 28877--28888, 2021.

\end{thebibliography}
